\pdfoutput=1

\documentclass[11pt]{article}

\usepackage{ACL2023}

\usepackage{times}
\usepackage{tcolorbox}
\usepackage{latexsym}
\usepackage{subcaption}
\usepackage{musicography}
\usepackage{arydshln}

\usepackage[T1]{fontenc}

\usepackage[utf8]{inputenc}

\usepackage{microtype}
\usepackage{enumitem}

\usepackage{inconsolata}
\usepackage{amsmath}
\usepackage{graphicx}
\usepackage{multirow}

\title{Same Company, Same Signal:\\ The Role of Identity in Earnings Call Transcripts}

\author{Ding Yu, Zhuo Liu,  Hangfeng He\\
University of Rochester\\
{\{ding.yu, zhuo.liu,  hangfeng.he\}@rochester.edu}
}

\begin{document}
\maketitle 
\begin{abstract}

Post-earnings volatility prediction is critical for investors, with previous works often leveraging earnings call transcripts under the assumption that their rich semantics contribute significantly. To further investigate how transcripts impact volatility, we introduce DEC, a dataset featuring accurate volatility calculations enabled by the previously overlooked \texttt{beforeAfterMarket} attribute and dense ticker coverage. Unlike established benchmarks, where each ticker has only around two earnings, DEC provides 20 earnings records per ticker. Using DEC, we reveal that post-earnings volatility undergoes significant shifts, with each ticker displaying a distinct volatility distribution. To leverage historical post-earnings volatility and capture ticker-specific patterns, we propose two training-free baselines: \textit{Post-earnings Volatility} (PEV) and \textit{Same-ticker Post-earnings Volatility} (STPEV). These baselines surpass all transcripts-based models on DEC as well as on established benchmarks. Additionally, we demonstrate that current transcript representations predominantly capture ticker identity rather than offering financially meaningful insights specific to each earnings. This is evidenced by two key observations: earnings representations from the same ticker exhibit significantly higher similarity compared to those from different tickers, and predictions from transcript-based models show strong correlations with prior post-earnings volatility\footnote{Our code is publicly available at \url{https://github.com/piqueyd/Same-Company-Same-Signal}}.
 

\end{abstract}

\section{Introduction}
Post-earnings volatility prediction is crucial for investors and an emerging trend in the field of financial natural language processing (FinNLP). Volatility, defined as the standard deviation of returns over a specific period—post earnings call in this context—is a key financial metric for evaluating a company’s performance. 


Traditional finance methods primarily rely on volatility time series and statistical techniques such as GARCH and its variants \cite{engle1982autoregressive, BOLLERSLEV1986307}. 
However, with the rapid advancements in natural language processing (NLP) and audio processing, numerous studies have focused on utilizing unstructured earnings call data, such as transcripts and audio recordings, to enhance post-earnings volatility prediction \cite{qin-yang-2019-say, yang2020html, CIKM2020MAEC}. In this pursuit, researchers have employed a variety of techniques, including heterogeneous graphs \cite{sawhney-etal-2020-voltage, Liu_Zhu_Wang_Ma_Yin_Zheng_2024}, language model pre-training \cite{yang2022numhtml, niu-etal-2023-kefvp} and Large Language Models (LLMs) \cite{ cao2024risklabs, cao2024ecc},  to better address this complex problem.


Delving deeper into the background of earnings calls, we found that previous benchmarks, EC \cite{qin-yang-2019-say} and MAEC \cite{CIKM2020MAEC}, have overlooked a crucial attribute: \texttt{beforeAfterMarket}, which indicates whether earnings are released before the market opens or after it closes. This attribute is indispensable for accurately calculating volatility. We also observed that EC and MAEC prioritize ticker\footnote{In this work, “ticker” refers to a company.} coverage breadth over density, as each ticker appears around twice in these datasets. This limitation prevents tracking a company’s earnings over the long term. To address this, we curated a dense earnings call dataset, DEC, where each ticker is represented with 20 earnings records. This enables robust long-term trend analysis and detailed quarter-to-quarter comparisons.

On DEC, we observe that post-earnings absolute returns—and consequently, volatility\footnote{The volatility calculation is detailed in Section~\ref{Volatility}.}—are significantly higher than during normal periods, and that each ticker exhibits distinct post-earnings volatility patterns, we thus hypothesize that historical post-earnings volatility plays a dominant role in volatility prediction. To this end, we introduce two training-free baselines: PEV \textit{(Post-earnings Volatility)} and STPEV \textit{(Same-ticker Post-earnings Volatility)}. Remarkably, even with a simple mean-based implementation, our approach achieves state-of-the-art (SOTA) performance on all datasets: EC, MAEC and DEC, compared to transcripts-based models. Through further comparisons at both the representation level and the prediction level, we find that transcripts from the same company exhibit high similarity, and the predictions of transcript-based models strongly correlate with those of STPEV(Mean).  
This suggests that \textit{transcripts primarily reflect ticker identity and prior post-earnings volatility distribution}, challenging the mainstream assumption that each earnings call provides financially meaningful semantics.

Our contributions include:
\begin{itemize}
    \item We curated a dense dataset, DEC, where each ticker includes 20 earnings, in contrast to the approximately two earnings per ticker in established datasets. Additionally, DEC incorporates the previously omitted \texttt{beforeAfterMarket} attribute, enabling accurate volatility calculations.
    \item 
    We propose two training-free baselines, PEV and STPEV, which achieve SOTA performance on EC, MAEC, and DEC, surpassing various transcripts-based models. 
    \item Through representation-level comparisons between examples from the same company and those across all companies, as well as prediction-level comparisons between STPEV and transcript-based models, we find that: \textbf{\textit{transcripts mostly reflect ticker identity}}.
\end{itemize}

\section{Related Work}
Considerable research efforts have been dedicated to leveraging earnings call transcripts, often in combination with other modalities such as audio recordings or time-series data, to model financial risk.


\paragraph{Transcripts-based models.} A few models rely exclusively on transcripts for volatility prediction. For instance, the Multi-Round QA Attention model \cite{ijcai2020p631} extracts semantic information from each question-answer round and integrates features across multiple granularities to predict volatility.

Transcripts are often combined with audio recordings during earnings calls. The Multimodal Deep Regression Model (MDRM) \cite{qin-yang-2019-say} integrates transcript and audio information to forecast volatility. Building on MDRM, the Hierarchical Transformer-based Model (HTML) \cite{yang2020html} employs a hierarchical transformer framework to enhance performance. Addressing the limitations of traditional language models in processing numerical information, which is critical in transcripts, Numerical HTML (NumHTML) \cite{yang2022numhtml} was developed. NumHTML improves predictive accuracy by incorporating numerical data, leveraging different categories of numbers and their magnitudes to augment the textual model's efficacy. 

Additionally, VolTAGE \cite{sawhney-etal-2020-voltage} and ECHO-GL \cite{Liu_Zhu_Wang_Ma_Yin_Zheng_2024} demonstrate that correlations between stocks are beneficial for predicting volatility. These models derive stock relationships from the rich semantic content of earnings calls using a heterogeneous graph learning.

In the era of LLM, RiskLabs \cite{cao2024risklabs} utilizes LLMs to encode transcripts and news articles, combining these with other modalities to deliver a comprehensive approach for volatility prediction. The ECC Analyzer \cite{cao2024ecc} employs LLMs to first extract paragraph-level general information by summarizing the text and subsequently identifies fine-grained focus sentences using Retrieval-Augmented Generation (RAG).

\citet{liu-etal-2024-beyond} highlight that existing pre-trained embedding models and LLM embeddings often fail to capture subtle shifts in financial narratives for the same company across different periods. To address this limitation, a specifically tailored LLM-augmented pipeline has been developed for financial semantic textual similarity.

In this work, we use different LLM embeddings to represent both vanilla transcripts and LLM fine-grained transcripts, which provide deeper insights and a more nuanced understanding.  



\paragraph{Time series-based models}
The Knowledge-enhanced Financial Volatility Prediction (KeFVP) model \cite{niu-etal-2023-kefvp} demonstrates the advantages of integrating time-series data with textual information. \textit{Pre-earnings volatility series},  is processed by the Autoformer \cite{wu2021autoformer}. The resulting representations are then conditionally combined with transcript representations, enhancing the model’s predictive capabilities. 

In this work, we introduce a second type of volatility series: \textit{the post-earnings volatility series}, which captures the sequence of volatility observed after a prior earnings announcement and before the subsequent one. In contrast, \textit{the pre-earnings volatility series} refers to the sequence of volatility recorded during the period leading up to a specific earnings announcement.


\section{Post Earnings Volatility Prediction}

\begin{figure*}[ht]
    \centering    \includegraphics[width=0.8\textwidth]{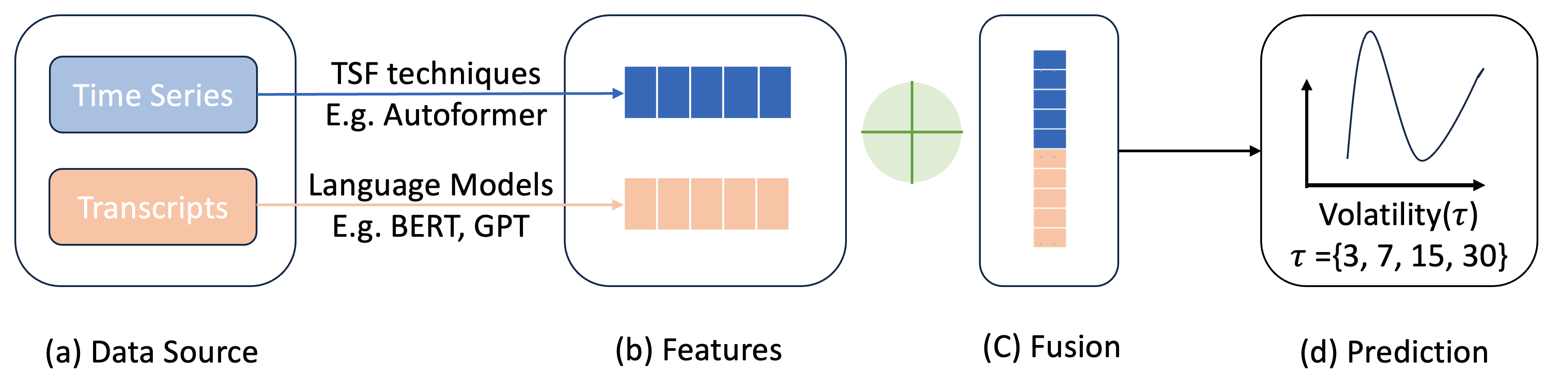}
    \caption{Overview of Post-Earnings Volatility Prediction: Time-series data is processed using time-series forecasting (TSF) techniques such as Autoformer \cite{wu2021autoformer}, while textual data is handled by language models like BERT \cite{devlin2018bert} and GPT \cite{radford2018improving}. The resulting representations are then combined for prediction.}
    \label{Figure: Overview}
\end{figure*}

\paragraph{Earnings call transcripts} Earnings call transcripts are written records of the earnings calls held by companies at the end of each quarter or fiscal year. These transcripts capture the detailed discussions about financial results, company performance, and future projections provided by the company’s executives, as well as the question and answer session with analysts and investors. 

\paragraph{Volatility}
\label{Volatility}
In financial terms, volatility \cite{kogan-etal-2009-predicting} represents the degree of variation of a trading price series over time as measured by the standard deviation of  returns. Let's assume earnings is released at day  \( t \), and mathematically, volatility can be defined over a specific interval, \([t, t+\tau]\), as follows:
\begin{equation}
\scriptstyle
v_{[t,t+\tau]} = \log {\sqrt{\sum_{i=0}^{\tau} (r_{t+i} - \overline{r})^2}},
\label{eq: volatility}   
\end{equation}
where \( r_{t+i} \) represents the return at time \( t+i \), calculated as: ${\scriptstyle r_{t+i} = \frac{C_{t+i} - C_{t+i-1}}{C_{t+i-1}}}$, and \( C_{i} \) denotes the closing price on day \( i \). Additionally, \( \overline{r} \) is the average return over the period from \( t \) to \( t+\tau \).

Volatility is a critical measure in finance as it reflects the risk associated with the price movements of a security. In previous work, \citet{qin-yang-2019-say} utilized various time intervals, \(\tau = \{3, 7, 15, 30\}\), to quantify volatility, capturing both short-term and long-term market behaviors. 


\paragraph{Volatility Time-Series}
The \textit{pre-earnings volatility time-series} represents volatility from day \( i-\tau+1 \) to day \( i-1 \), while the \textit{post-earnings volatility time-series} represents volatility for the \( i-\tau+1 \)th to the \( i-1 \)th earnings. The former reflects volatility leading up to an earnings announcement, while the latter captures volatility occurring between a prior and subsequent earnings announcement.

Figure~\ref{Figure: Overview} illustrates common approaches that leverage transcripts and volatility time series. These are processed separately, and the resulting representations are then combined for prediction.

\section{A Dense Dataset: DEC}

\subsection{Two Widely-Used Datasets}
For the task of post-earnings volatility prediction, two datasets are commonly employed: EC \cite{qin-yang-2019-say} and MAEC \cite{CIKM2020MAEC}. The MAEC dataset is further divided into two subsets, corresponding to the years 2015 and 2016.

\subsection{Missing BeforeAfterMarket}
\label{BeforeAfterMarket}
During our analysis of the EC and MAEC datasets, we identified a critical oversight regarding the timing of earnings releases—specifically, the \texttt{beforeAfterMarket} attribute, which indicates whether the release occurs before or after the market opens. This attribute is essential for accurately calculating post-earnings volatility. 
Previous volatility calculations \footnote{\url{https://github.com/hankniu01/KeFVP/tree/main/price\_data}} fail to account for scenarios \textit{where earnings are released before the market opens}, requiring the current trading day to be treated as the starting point of the volatility calculation period. A specific example is illustrated in Section~\ref{Appendix Important BeforeAfterMarket}. Unfortunately, previous studies have often neglected this critical factor, resulting in inaccuracies in volatility measurements.



\subsection{Sparse Same-ticker Representation}
Tracking a company’s earnings call transcripts over the long term provides valuable insights for investors by offering a deeper understanding of the company’s management, strategy, and market positioning.  It also reveals key performance metrics, progress on stated goals, and emerging trends that could influence future growth. Comparing transcripts over time allows investors to detect shifts in narratives or priorities, which may signal potential challenges or opportunities. 

To investigate whether the current datasets supports long-termism, we define the Overlapping Earnings per Ticker (OET) as follows:
\begin{equation}
    \scriptstyle
    \text{OET} = \frac{\text{Count (testing tickers overlapped training earnings)}} {\text{Count (testing tickers)}}
    \label{eq:rot_formula}
\end{equation}

This metric measures how well the training set\footnote{Here, we include both the training set and validation set.} represents the testing set in terms of \textbf{ticker overlap}. Generally, a higher OET value indicates a longer-term focus on the tickers' earnings.

Table~\ref{table: EC and MAEC Dataset Statistics} shows the statistics for the EC and MAEC datasets. Regrettably, all three datasets exhibit low OET values, indicating a limited number of overlapping earnings. This limitation prevents investors from tracking a company’s long-term earnings performance, which provides valuable insights into growth trends, financial health, and management effectiveness.


\begin{table}[ht]
\centering
\resizebox{\columnwidth}{!}{%
\begin{tabular}{cc|ccc|cc}
\hline
Dataset & Cate & \# E  & \# T & Ratio $\frac{\#E}{\#T}$ & \# OE & OET \\
\hline
\multirow{4}{*}{EC} 
 & All & 559 & 272 & 2.055 & - & - \\
 & Train & 391 & 243 & 1.609 & - & - \\
 & Val & 56 & 56 & 1.0 & - & - \\
 & Test & 112 & 112 & 1.0 & 178 & 1.589\\
\hline
\multirow{4}{*}{MAEC-15} 
& All & 765 & 527 & 1.452 & - & - \\ 
& Train & 535 & 409 & 1.308 & - & - \\ 
& Val & 76 & 76 & 1.0 & - & - \\ 
& Test & 154 & 154 & 1.0 & 94 & 0.61 \\  
\hline
\multirow{4}{*}{MAEC-16} 
& All & 1400 & 908 & 1.542 & - & -  \\
& Train & 980 & 734 & 1.335 & - & -  \\ 
& Val & 140 & 140 & 1.0 & - & -  \\
& Test & 280 & 277 & 1.011 & 215 & 0.768 \\
\hline
\end{tabular}%

}
\caption{Statistics for EC, MAEC-15, and MAEC-16. \# E , \# T and \# OE defines the number of earnings, tickers, and overlapping earnings respectively. OET is the Overlapping Earnings per Ticker defined in equation~\ref{eq:rot_formula}.}
\label{table: EC and MAEC Dataset Statistics}
\end{table}

\subsection{A Dense Dataset: DEC}
To address these limitations, we curated a new dataset: \textbf{DEC}. The DEC dataset offers four key advantages over the existing datasets:

\begin{itemize}
    \item \textbf{Correct Volatility Calculation:} As described in Section~\ref{BeforeAfterMarket}, the \texttt{beforeAfterMarket} attribute is omitted in existing datasets. To address this, we collect this important attribute from the financial data provider EOD: \footnote{ \url{https://eodhd.com/}} to ensure accurate volatility calculation.
    \item \textbf{Longitudinal Depth:} DEC comprises 1,800 earnings, providing a temporally dense focus on 90 specific tickers over \textit{20} quarters, spanning the period from 2019 to 2023.
    

   \item \textbf{Latitudinal Depth:} The dataset includes representative tickers from various sectors\footnote{\url{https://seekingalpha.com/etfs-and-funds/etf-tables/key_markets}} within the U.S. market, ensuring representation across a diverse range of industries.
    
    
    \item \textbf{Recency and Relevance:} DEC is more recent compared to existing datasets, offering up-to-date information and reflecting the latest market dynamics. Notably, it also encompasses the COVID-19 pandemic period, which triggered a corporate finance crisis and resulted in distinct patterns compared to typical market conditions \cite{ellul2020covid}. 
\end{itemize}

Table~\ref{table:DEC Dataset Statistics} presents the OET statistics of the DEC dataset. It is evident that the OET values increase over time, as indicated by the progression from the top-left to the bottom-right of the table. Further details regarding the curation process of the DEC dataset can be found in Appendix~\ref{Appendix Dataset DEC}. With its dense ticker coverage, DEC enables long-term trend analysis and quarter-to-quarter analysis by comparing sequential performance, and understanding transitions between quarters.



\begin{table*}[ht]
\centering
\scriptsize
\resizebox{\textwidth}{!}{%
\begin{tabular}{|c|c c c|c c c|c c c|c c c|}
\hline
\multirow{3}{*}{Year} &
\multicolumn{3}{c|}{First Quarter} &
\multicolumn{3}{c|}{Second Quarter} &
\multicolumn{3}{c|}{Third Quarter} &
\multicolumn{3}{c|}{Fourth Quarter}
\\

\hline
& Count(Training) &  Count(Testing) & OET
& Count(Training) &  Count(Testing) & OET
& Count(Training) &  Count(Testing) & OET
& Count(Training) &  Count(Testing) & OET
\\

\hline
2019 & 0 & 90 & 0 & 90 & 90 & 1.0 & 180 & 90 & 2.0 & 270 & 90 & 3.0 \\
\hline
2020 & 360 & 90 & 4.0 & 450 & 90 & 5.0 & 540 & 90 & 6.0 & 630 & 90 & 7.0 \\
\hline
2021 & 720 & 90 & 8.0 & 810 & 90 & 9.0 & 900 & 90 & 10.0 & 990 & 90 & 11.0 \\
\hline
2022 & 1080 & 90 & 12.0 & 1170 & 90 & 13.0 & 1260 & 90 & 14.0 & 1350 & 90 & 15.0 \\
\hline
2023 & 1440 & 90 & 16.0 & 1530 & 90 & 17.0 & 1620 & 90 & 18.0 & 1710 & 90 & 19.0 \\
\hline

\end{tabular}
}
\caption{DEC Dataset Statistics. The dataset focuses on 90 tickers in the U.S. market, spanning 20 quarters.  It is \textit{dense in ticker coverage} and OET values, defined in equation~\ref{eq:rot_formula}, increase
over time.}
\label{table:DEC Dataset Statistics}
\end{table*}

\begin{figure*}[!htb]  
    \centering

    \begin{subfigure}{.24\textwidth}
        \centering
        \includegraphics[width=\linewidth]{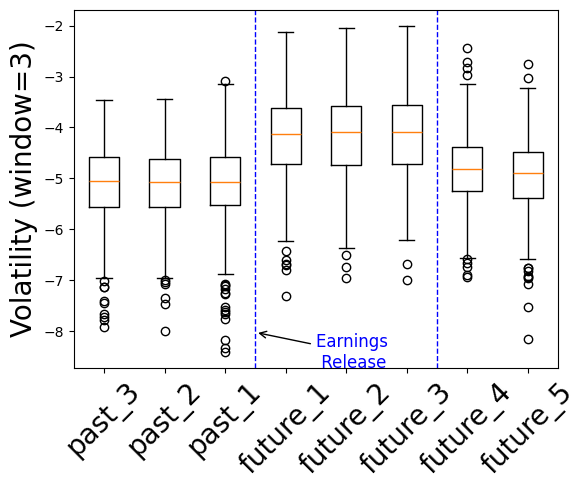}
        \caption{EC}
        \label{EC E V.S. NE}
    \end{subfigure}%
    \hfill 
    \begin{subfigure}{.24\textwidth}
        \centering
        \includegraphics[width=\linewidth]{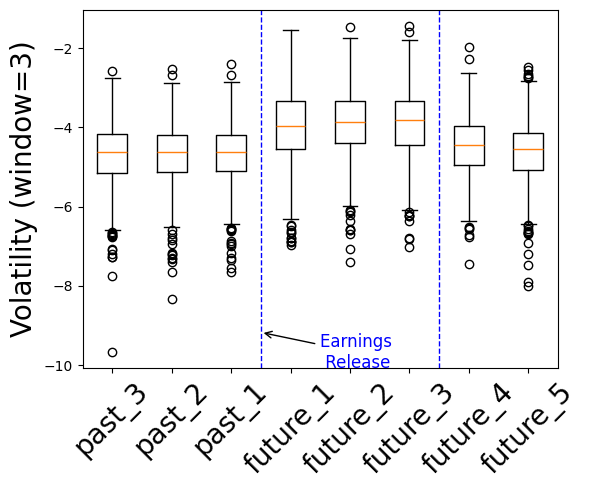}
        \caption{MAEC-15}
        \label{MAEC-15 E V.S. NE}
    \end{subfigure}%
    \hfill 
    \begin{subfigure}{.24\textwidth}
        \centering
        \includegraphics[width=\linewidth]{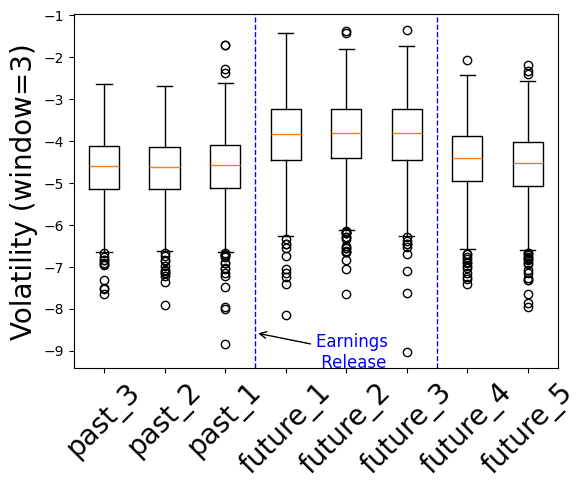}
        \caption{MAEC-16}
        \label{MAEC-16 E V.S. NE}
    \end{subfigure}
    \begin{subfigure}{.24\textwidth}
        \centering
        \includegraphics[width=\linewidth]{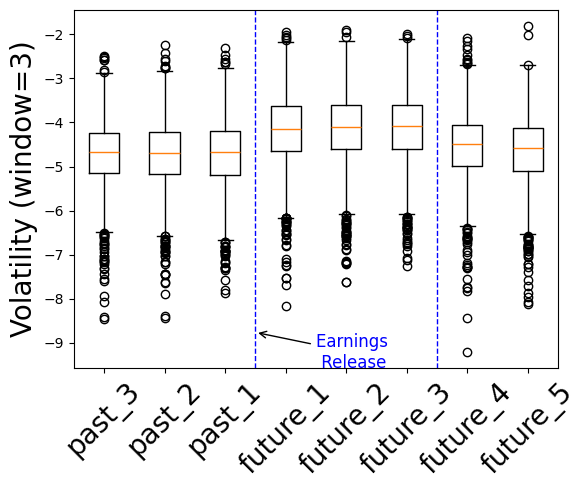}
        \caption{DEC}
        \label{DEC E V.S. NE}
    \end{subfigure}
    \caption{Comparison of three-day volatility before and after earnings announcements. Earnings are released between the day labeled \texttt{past\_1} and the day labeled \texttt{future\_1}. Days where the volatility calculation involves the return of \texttt{future\_1} exhibit significantly higher volatility compared to others. This pattern holds consistently across all windows \{3, 7, 15, 30\}. Further details are provided in Appendix~\ref{Appendix Post Earnings Volatility Distribution Drift}.}
\label{Figure: Volatility Comparison Before V.S. After E}
\end{figure*}

\section{On the Importance of Past Post-earnings Volatility}

\subsection{Distribution Shift After Earnings}
\label{Distribution Shift After Earnings}
We observe a significant increase in absolute returns following earnings announcements across the EC, MAEC, and DEC datasets. Specifically, the absolute return on the first day after earnings, \( r_{future_1} \), is consistently higher than on pre-earnings days or other subsequent post-earnings days. We hypothesize that this phenomenon stems from the market’s reaction to freshly disclosed and potentially unexpected financial information from the company. Details are in Appendix~\ref{Appendix Post Earnings Volatility Distribution Drift}.


According to the definition of volatility in equation~\ref{eq: volatility}, we conclude that volatility calculations involving the first daily return after earnings, \( r_{future_1} \), should be higher compared to periods without it. Given \( \tau \) as the volatility window, a total of \( \tau \) days of volatility are directly influenced by \( r_{future_1} \). As illustrated in Figure~\ref{Figure: Volatility Comparison Before V.S. After E}, which compares three-day volatility before and after earnings announcements across the EC, MAEC and DEC datasets\footnote{The \texttt{beforeAfterMarket} attribute is adjusted based on the original EC and MAEC datasets. The plot without \texttt{beforeAfterMarket} adjustment is provided in Appendix~\ref{Appendix Post Earnings Volatility Distribution Drift}}, post-earnings volatility within a three-day window is notably higher than the volatility observed on other trading days. This pattern remains consistent across other time windows (7, 15, and 30 days), with further details in Appendix~\ref{Appendix Post Earnings Volatility Distribution Drift}.

Given the pronounced differences in volatility distribution between post-earnings periods\footnote{Precisely, post-earnings periods here refer to the days where calculations involving the first return after earnings.} and non-earnings periods, we hypothesis that:

\begin{center}
\textbf{\textit{Incorporating \underline{past} post-earnings volatility is essential for volatility prediction.}}
\end{center}                    


\subsection{Ticker-Specific Volatility Signature}
\label{Ticker-Specific Volatility Regime}

\begin{figure}[ht]
    \centering    \includegraphics[width=0.46\textwidth]{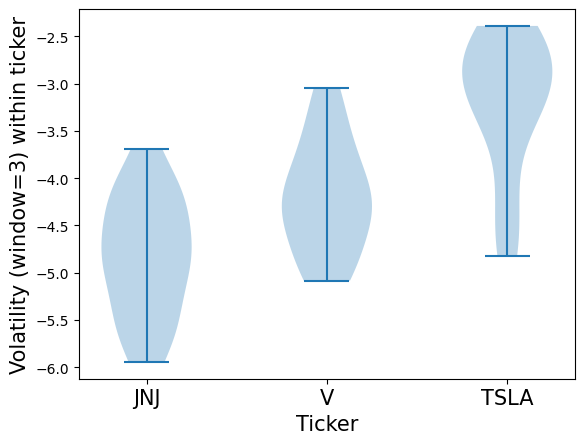}
    \caption{Three-day post earnings volatility comparison across companies: JNJ(Johnson \& Johnson), V(Visa), and TSLA(Tesla). Totally 20 earnings, from 2019 to 2023, are involved for plot.}
    \label{Figure: Ticker-Specific PE Patterns}
\end{figure}

It also has been observed that the post-earnings volatility for each company tends to follow a distinct distribution. As depicted in Figure~\ref{Figure: Ticker-Specific PE Patterns}, companies such as JNJ, V, and TSLA\footnote{JNJ, V, and TSLA refer to Johnson \& Johnson, Visa, and Tesla, respectively.} exhibit markedly different three-day post-earnings volatility patterns. 

We term this phenomenon as \textit{Volatility Signature}, reflecting the persistence of a company’s volatility patterns over time. This signature likely arises from intrinsic company characteristics that remain relatively stable over short periods. These characteristics may include industry and sector classification, operational dynamics, company size and market position, and financial structure. Motivated by the \textit{Volatility Signature}, we refine our hypothesis:


\begin{center}
\textbf{\textit{Incorporating \underline{same-ticker} past post-earnings volatility is critical for volatility prediction.}}
\end{center}

\begin{table*}[ht]
\centering
\resizebox{\textwidth}{!}{%
\begin{tabular}{|c|c c c c c|c c c c c|c c c c c|c|}
\hline
\multirow{2}{*}{Model} & 
\multicolumn{5}{c|}{EC} &
\multicolumn{5}{c|}{MAEC-15} &
\multicolumn{5}{c|}{MAEC-16} &
\multirow{2}{*}{Average} \\
& $\overline{MSE}$ & $MSE_3$ & $MSE_7$ & $MSE_{15}$ & $MSE_{30}$ 
& $\overline{MSE}$ & $MSE_3$ & $MSE_7$ & $MSE_{15}$ & $MSE_{30}$
& $\overline{MSE}$ & $MSE_3$ & $MSE_7$ & $MSE_{15}$ & $MSE_{30}$ & \\

\hline
Vpast & 1.12 & 2.99 & 0.83 & 0.42 & 0.23 & - & - & - & - & - & - & - & - & - & - & - \\
Price LSTM & 0.75 & 1.97 & 0.46 & 0.32 & 0.24 & - & - & - & - & - & - & - & - & - & -& - \\
BiLSTM + ATT & 0.74 & 1.98 & 0.44 & 0.30 & 0.23 & 0.696 & 1.599$^\sharp$ & 0.560$^\sharp$ & 0.339$^\sharp$ & 0.284$^\sharp$ & 0.691 & 1.544$^\sharp$ & 0.571$^\sharp$ & 0.362$^\sharp$ & 0.288$^\sharp$ & 0.709 \\
HAN(Glove) & 0.60 & 1.43 & 0.46 & 0.31 & 0.20 & - & - & - & - & - & - & - & - & - & - & - \\
MDRM(Audio) & 0.60 & 1.41 & 0.44 & 0.32 & 0.22 & - & - & - & - & - & - & - & - & - & - & - \\
MDRM(Text+Audio) & 0.58 & 1.37 & 0.42 & 0.30 & 0.22 & 0.630 & 1.425$^\sharp$ & 0.488$^\sharp$ & 0.320$^\sharp$ & 0.285$^\sharp$ & 0.618 & 1.426$^\sharp$ & 0.476$^\sharp$ & 0.311$^\sharp$ & 0.259$^\sharp$ & 0.609 \\
HTML(Text) & 0.46 & 1.18 & 0.37 & 0.15 & 0.13 & 0.514 & 1.199$^\sharp$ & 0.440$^\sharp$ & 0.231$^\sharp$ & 0.187$^\sharp$ & 0.579 & 1.287$^\sharp$ & 0.479$^\sharp$ & 0.300 & 0.249$^\sharp$ & 0.518\\
HTML(Text+Audio) & 0.40 & 0.85 & 0.35 & 0.25 & 0.16 & 0.487 & 1.065$^\sharp$ & 0.416$^\sharp$ & 0.272$^\sharp$ & 0.196$^\sharp$ & 0.556 & 1.160$^\sharp$ & 0.515$^\sharp$ & 0.314$^\sharp$ & 0.236$^\sharp$ & 0.481\\

VolTAGE & 0.31 & 0.63 & 0.29 & 0.17 & 0.14 & - & - & - & - & - & - & - & - & - & - & - \\
KeFVP$^{\musDoubleFlat}$ & \underline{0.300} & 0.610 & 0.291 & 0.183 & 0.114 & \textbf{0.204} & 0.418 & 0.187 & 0.122 & 0.087 & {0.318} & 0.445 & 0.279 & 0.303 & 0.177 & 0.274 \\
\hline

SVM(TF-IDF)$^b$ & 0.70 & 1.70 & 0.50 & 0.34 & 0.25 & - & - & - & - & - & - & - & - & - & - & -\\
bc-LSTM$^b$ & 0.59 & 1.42 & 0.44 & 0.30 & 0.22 & - & - & - & - & - & - & - & - & - & -  & - \\
Multi-Fusion CNN$^b$ & 0.41 & 0.73 & 0.35 & 0.29 & 0.28 & - & - & - & - & - & - & - & - & - & -  & - \\
NumHTML(Text+Audio)$^\natural$ & 0.31 & - & - & - & - & - & - & - & - & - & - & - & - & - & -  & - \\
Ensemble(Text+Audio)$^b$ & 0.302 & 0.601 & 0.308 & 0.181 & 0.119 & - & - & - & - & - & - & - & - & - & -  & - \\
RiskLabs$^{\dagger}$ & 0.324 & 0.585 & 0.317 & 0.233 & 0.171 & - & - & - & - & - & - & - & - & - & -  & - \\

ECC Analyzer $^{\ddagger}$ & 0.314 & 0.553 & 0.306 & 0.237 & 0.158 & - & - & - & - & - & - & - & - & - & -  & - \\

\hline\hline

{GPT4o Pred (3 shot)}  $^{\alpha}$ & 0.609 & 1.433 & 0.501 & 0.26 & 0.244 & 0.345 & 0.585 & 0.404 & 0.22 & 0.169 & 0.441 & 0.545 & 0.538 & 0.51 & 0.171 & 0.465 \\

{Gemini Pred (3 shot)}  $^{\alpha}$ & 0.592 & 1.368 & 0.487 & 0.251 & 0.263 & 0.451 & 0.824 & 0.598 & 0.219 & 0.163 &  0.337 & 0.475 & 0.404 & 0.3 & 0.171 & 0.466\\

{Vanilla(Voyage)} $^{\alpha}$  & 0.387 & 0.751 & 0.375 & 0.245 & 0.177 & 0.34 & 0.623 & 0.303 & 0.232 & 0.201 & 0.274 & 0.458 & 0.267 & 0.222 & 0.151 & 0.334 \\ 
{Vanilla(Gecko)} $^{\alpha}$ & 0.36 & 0.664 & 0.356 & 0.237 & 0.182 & 0.283 & 0.513 & 0.27 & 0.19 & 0.159 & 0.254 & 0.402 & 0.25 & 0.249 & 0.116 & 0.299 \\ 
{Vanilla(OpenAI)} $^{\alpha}$ & 0.339 & 0.682 & 0.311 & 0.209 & 0.155 &
0.319 & 0.596 & 0.290 & 0.206 & 0.184 & 
\underline{0.235} & 0.402 & 0.232 & 0.177 & 0.127 & 0.298 \\

{GPT4o(Summarization)} $^{\alpha}$ & 0.299 & 0.585 & 0.283 & 0.188 & 0.142 & 0.3 & 0.548 & 0.274 & 0.204 & 0.174 &  0.246 & 0.405 & 0.236 & 0.221 & 0.124 & 0.282 \\ 
{GPT4o(Task-Specific)} $^{\alpha}$ & 0.314 & 0.624 & 0.294 & 0.195 & 0.142 &  0.268 & 0.505 & 0.248 & 0.164 & 0.155 & 0.232 & 0.372 & 0.232 & 0.215 & 0.111 & 0.271 \\
{Gemini(Summarization)} $^{\alpha}$ & 0.275 & 0.568 & 0.244 & 0.167 & 0.122 & 0.268 & 0.494 & 0.254 & 0.165 & 0.158 & 0.23 & 0.372 & 0.236 & 0.202 & 0.109 & \underline{0.258} \\
{Gemini(Task-Specific)} $^{\alpha}$ & 0.284 & 0.583 & 0.252 & 0.175 & 0.128 & 0.276 & 0.508 & 0.262 & 0.176 & 0.158 & 0.235 & 0.383 & 0.229 & 0.212 & 0.115 & 0.265 \\

\hline

\textbf{PEV(Mean)} $^{\alpha}$ & 0.399 & 0.743 & 0.389 & 0.262 & 0.201 & 
0.305 & 0.532 & 0.301 & 0.209 & 0.177 &
\underline{0.23} & 0.38 & 0.229 & 0.173 & 0.139 & 0.311 \\

\textbf{STPEV(Mean)} $^{\alpha}$ & 0.349 & 0.724 & 0.33 & 0.205 & 0.138 &
0.301 & 0.571 & 0.273 & 0.19 & 0.17 & 
0.271 & 0.459 & 0.273 & 0.209 & 0.144 & 0.307 \\

\textbf{PEV(Mean)(Aug)} $^{\alpha}$ & 0.367 & 0.712 & 0.351 & 0.235 & 0.17 & 
0.283 & 0.514 & 0.271 & 0.188 & 0.157 & 
\textbf{0.229 }& 0.37 & 0.24 & 0.168 & 0.139 & 0.293 \\
\textbf{STPEV(Mean)(Aug)} $^{\alpha}$ & \textbf{0.296} & 0.569 & 0.293 & 0.201 & 0.122 & 
\underline{0.225} & 0.443 & 0.214 & 0.13 & 0.112 &
0.25 & 0.5 & 0.237 & 0.159 & 0.103 & \textbf{0.257} \\
\hline

\end{tabular}
}

\caption{The overall performance on EC, MAEC-15 and MAEC-16 datasets.The models below the double line and marked with ${\alpha}$ are implemented in this work. The results with $\natural$, $\sharp$, $b$, ${\musDoubleFlat}$, ${\dagger}$ and ${\ddagger}$ are retrieved from \citet{yang2022numhtml}, \citet{CIKM2020MAEC}, \citet{10.1145/3394171.3413752}, \citet{niu-etal-2023-kefvp}, \citet{cao2024risklabs}
and \citet{cao2024ecc} respectively, and the remainder are from \citet{sawhney-etal-2020-voltage}. The best results are in bold, and the second-best results are underlined. To ensure a fair comparison, the \texttt{beforeAfterMarket} is not adjusted for the results presented here.}
\label{table:The Overall Performance}
\end{table*}

\subsection{Simple Baselines}

\paragraph{PEV} Building on our analysis, we propose a training-free baseline for post-earnings volatility prediction, referred to as the \textit{Post-earnings Volatility} (PEV) model. The PEV(X) model primarily utilizes \textit{historical post-earnings volatility} as input and applies an aggregation function, \( X \), which we instantiate as the mean in this work.

\paragraph{STPEV} To incorporate the concept of \textit{Volatility Signature}, we introduce a refined variant of the PEV, termed the \textit{Same-ticker Post-earnings Volatility} (STPEV). This variant exclusively leverages \textit{historical post-earnings volatility from the same ticker}, thereby improving its capacity to capture the characteristics specific to each company.




\section{Main Results}
\label{Main Results}

\subsection{Evaluations on EC and MAEC Datasets}
\label{Evaluations on EC and MAEC Datasets}

\paragraph{Augmentation on EC and MAEC datasets.}
As shown in Table~\ref{table: EC and MAEC Dataset Statistics}, EC, MAEC-15 and MAEC-16 datasets all suffer from a limited number of same-ticker earnings, which restricts STPEV’s ability to fully exploit the ticker-specific volatility signature. To mitigate this limitation, we augment these datasets by extending their historical range from the past 5 years.\footnote{We only extend the price records, such as close price, daily return, and volatility, without the earnings call transcripts.}. Appendix~\ref{Appendix Augmentation on EC and MAEC} contrasts the data statistics between the original and augmented datasets. By augmenting the datasets, we achieve  higher OET values, enhancing the suitability of the datasets for both PEV and STPEV.

\paragraph{Baselines}
\label{Baselines}
To validate the effectiveness of PEV and STPEV, we benchmark their performance against several established methods. These baseline methods include MDRM (Text+Audio) \cite{qin-yang-2019-say}, HTML (Text+Audio) \cite{yang2020html}, VolTAGE \cite{sawhney-etal-2020-voltage}, NumHTML (Text+Audio) \cite{yang2022numhtml}, KeFVP \cite{niu-etal-2023-kefvp}, RiskLabs \cite{cao2024risklabs}, and ECC Analyzer \cite{cao2024ecc}.

In addition to previous works, we also implement several transcripts-based baselines:

\begin{itemize}
    \item \textbf{LLM Direct Prediction:} We prompt the LLMs\footnote{GPT4o-2024-08-06 and Gemini-1.5-Flash} using few-shot learning with the task description and the previous \textit{(earnings call transcripts, volatility)} pairs to directly predict the volatility. Details are in Appendix~\ref{Appendix LLM Fine-grained Texts}.
    
    \item \textbf{Vanilla Text:} Earnings call transcripts are directly processed to generate text embeddings using various models, including OpenAI embeddings, Gecko embeddings \cite{lee2024geckoversatiletextembeddings}, and a financial-domain-specific model, Voyage embeddings\footnote{Specifically, OpenAI text-embedding-3-large model, Text-embedding-005 model, and Voyage-Finance-2 model.}. These embeddings are processed by a simple 2-layer MLP model, which is also used for LLM fine-grained text.
    
    \item \textbf{LLM Fine-Grained Text:} Earnings call transcripts are scrutinized using LLMs, specifically GPT-4o and Gemini-1.5-Flash, employing two distinct strategies: \textit{Summarization} and \textit{Task-Specific}, to generate fine-grained summaries, which are then used to obtain embeddings\footnote{OpenAI embeddings are used for fine-grained texts.}. Details, such as prompt templates, can be found in Appendix~\ref{Appendix LLM Fine-grained Texts}.

\end{itemize}

\paragraph{PEV and STPEV settings.}
We evaluate PEV and STPEV with the following two simple implementations: PEV(Mean) and STPEV(Mean). 

\paragraph{Analysis.}
As shown in Table~\ref{table:The Overall Performance}, direct predictions from LLMs exhibit the worst performance, with MSEs of 0.465 and 0.466 for GPT4o and Gemini, respectively, highlighting the limited ability of LLMs to effectively handle regression tasks.

The vanilla transcripts using OpenAI and Gecko embeddings outperform all previous works except KeFVP \cite{niu-etal-2023-kefvp}, demonstrating the superior capability of LLM-based embeddings. Utilizing LLMs to process and analyze transcripts further improves performance, achieving the best MSE of 0.258 with the Gemini(summarization) strategy.

The PEV(Mean) and STPEV(Mean) achieve MSEs of 0.311 and 0.307 on original datasets, which are reasonable given the scarcity of prior earnings, with details in Appendix~\ref{Appendix Augmentation on EC and MAEC}. However, when applying the PEV(Mean) and STPEV(Mean) to the augmented datasets, the MSEs decrease to 0.293 and 0.257, with the latter achieving SOTA.

\begin{table*}[ht]
\centering
\resizebox{\textwidth}{!}{%
\begin{tabular}{|c|c|c c c c c|c c c c c|c c c c c|c c c c c|c|}
\hline
\multirow{2}{*}{Year} &
\multirow{2}{*}{Model} & 
\multicolumn{5}{c|}{First Quarter} &
\multicolumn{5}{c|}{Second Quarter} &
\multicolumn{5}{c|}{Third Quarter} &
\multicolumn{5}{c|}{Fourth Quarter} &
\multirow{2}{*}{Average}
 
\\
& & $\overline{MSE}$ & $MSE_3$ & $MSE_7$ & $MSE_{15}$ & $MSE_{30}$ 
& $\overline{MSE}$ & $MSE_3$ & $MSE_7$ & $MSE_{15}$ & $MSE_{30}$
& $\overline{MSE}$ & $MSE_3$ & $MSE_7$ & $MSE_{15}$ & $MSE_{30}$ 
& $\overline{MSE}$ & $MSE_3$ & $MSE_7$ & $MSE_{15}$ & $MSE_{30}$ & \\

\hline 
\multirow{10}{*}{2021} & GPT4o Pred (8 shot) & 0.306 & 0.713 & 0.235 & 0.152 & 0.124 & 0.357 & 0.707 & 0.237 & 0.187 & 0.297 & 0.439 & 0.742 & 0.414 & 0.318 & 0.284 & 0.293 & 0.653 & 0.247 & 0.153 & 0.12 & 0.349 \\

& Gemini Pred (8 shot) & 0.273 & 0.583 & 0.243 & 0.146 & 0.121 & 0.327 & 0.563 & 0.223 & 0.204 & 0.317 & 0.471 & 0.972 & 0.44 & 0.289 & 0.181 & 0.347 & 0.696 & 0.329 & 0.199 & 0.163 & 0.354 \\

& Vanilla (OpenAI) & \underline{0.170} & 0.357 & 0.148 & 0.079 & 0.097 & 0.250 & 0.457 & 0.212 & 0.145 & 0.185 & 0.372 & 0.547 & 0.462 & 0.285 & 0.194 & \textbf{0.213} & 0.501 & 0.173 & 0.109 & 0.068 & 0.251 \\

& Vanilla (Gecko) & 0.200 & 0.419 & 0.191 & 0.104 & 0.087 & 0.269 & 0.464 & 0.245 & 0.176 & 0.189 & 0.350 & 0.523 & 0.377 & 0.291 & 0.211 & 0.253 & 0.535 & 0.223 & 0.161 & 0.094 & 0.268 \\

& GPT4o (Summarization) & 0.183 & 0.390 & 0.162 & 0.097 & 0.084 & 0.277 & 0.482 & 0.252 & 0.171 & 0.204 & 0.353 & 0.549 & 0.430 & 0.278 & 0.156 & 0.234 & 0.544 & 0.190 & 0.124 & 0.079 & 0.262 \\

& GPT4o (Task-specific) & 0.177 & 0.388 & 0.164 & 0.088 & 0.070 & \textbf{0.246} & 0.444 & 0.214 & 0.145 & 0.180 & 0.357 & 0.515 & 0.428 & 0.295 & 0.189 & 0.242 & 0.537 & 0.197 & 0.145 & 0.088 & 0.255 \\

& Gemini (Summarization) & 0.175 & 0.356 & 0.153 & 0.083 & 0.106 & \textbf{0.246} & 0.454 & 0.212 & 0.144 & 0.173 & \textbf{0.322} & 0.502 & 0.394 & 0.240 & 0.151 & 0.255 & 0.553 & 0.215 & 0.142 & 0.109 & \underline{0.249} \\

& Gemini (Task-specific) & 0.176 & 0.384 & 0.159 & 0.094 & 0.067 & 0.249 & 0.435 & 0.210 & 0.156 & 0.197 & 0.347 & 0.503 & 0.407 & 0.286 & 0.190 & 0.248 & 0.548 & 0.215 & 0.140 & 0.088 & 0.255 \\

& Random (All) & 0.249 & 0.472 & 0.231 & 0.150 & 0.143 & 0.294 & 0.497 & 0.291 & 0.189 & 0.201 & 0.433 & 0.603 & 0.512 & 0.359 & 0.259 & 0.300 & 0.577 & 0.280 & 0.212 & 0.132 & 0.319 \\

& Random (Ticker) & 0.190 & 0.380 & 0.171 & 0.112 & 0.096 & 0.275 & 0.449 & 0.246 & 0.180 & 0.226 & 0.381 & 0.555 & 0.414 & 0.321 & 0.236 & 0.255 & 0.535 & 0.224 & 0.163 & 0.097 & 0.275 \\

\cline{2-23}

& \textbf{PEV(Mean)} & 0.216 & 0.433 & 0.209 & 0.115 & 0.105 & 0.271 & 0.451 & 0.239 & 0.184 & 0.209 & 0.405 & 0.580 & 0.429 & 0.342 & 0.270 & 0.288 & 0.568 & 0.260 & 0.199 & 0.127 & 0.295 \\

& \textbf{STPEV(Mean)} & \textbf{0.156} & 0.368 & 0.149 & 0.067 & 0.041 & 0.249 & 0.463 & 0.209 & 0.150 & 0.173 & \underline{0.333} & 0.525 & 0.353 & 0.260 & 0.196 & \underline{0.222} & 0.536 & 0.177 & 0.114 & 0.062 & \textbf{0.240} \\

\hline\hline

\multirow{10}{*}{2022} & GPT4o Pred (8 shot) & 0.416 & 0.964 & 0.348 & 0.214 & 0.137 & 0.39 & 0.814 & 0.368 & 0.239 & 0.141 & 0.349 & 0.85 & 0.308 & 0.162 & 0.076 & 0.316 & 0.683 & 0.268 & 0.203 & 0.109 & 0.368 \\

& Gemini Pred (8 shot) & 0.467 & 1.013 & 0.43 & 0.235 & 0.19 & 0.326 & 0.682 & 0.283 & 0.212 & 0.127 & 0.34 & 0.808 & 0.299 & 0.154 & 0.099 & 0.297 & 0.69 & 0.248 & 0.17 & 0.078 & 0.357 \\

& Vanilla (OpenAI) & \textbf{0.258} & 0.523 & 0.253 & 0.160 & 0.095 & 0.354 & 0.671 & 0.251 & 0.260 & 0.235 & 0.261 & 0.622 & 0.237 & 0.112 & 0.072 & \underline{0.256} & 0.585 & 0.195 & 0.156 & 0.088 & 0.282 \\

& Vanilla (Gecko) & 0.302 & 0.603 & 0.293 & 0.188 & 0.125 & 0.353 & 0.659 & 0.325 & 0.238 & 0.188 & 0.265 & 0.608 & 0.234 & 0.134 & 0.083 & 0.274 & 0.588 & 0.228 & 0.178 & 0.101 & 0.298 \\

& GPT4o (Summarization) & 0.332 & 0.641 & 0.311 & 0.225 & 0.149 & 0.317 & 0.656 & 0.273 & 0.174 & 0.164 & \textbf{0.239} & 0.586 & 0.195 & 0.107 & 0.069 & 0.291 & 0.629 & 0.217 & 0.201 & 0.118 & 0.295 \\

& GPT4o (Task-specific) & 0.309 & 0.585 & 0.286 & 0.220 & 0.147 & \textbf{0.307} & 0.618 & 0.196 & 0.193 & 0.220 & \underline{0.243} & 0.578 & 0.219 & 0.110 & 0.065 & 0.275 & 0.599 & 0.209 & 0.194 & 0.099 & 0.284 \\

& Gemini (Summarization) & 0.297 & 0.591 & 0.269 & 0.208 & 0.119 & 0.348 & 0.660 & 0.240 & 0.244 & 0.246 & 0.252 & 0.605 & 0.220 & 0.109 & 0.073 & \textbf{0.250} & 0.573 & 0.180 & 0.151 & 0.094 & 0.287 \\

& Gemini (Task-specific) & 0.291 & 0.599 & 0.276 & 0.169 & 0.120 & 0.314 & 0.656 & 0.254 & 0.187 & 0.157 & \underline{0.243} & 0.584 & 0.216 & 0.109 & 0.065 & 0.267 & 0.604 & 0.191 & 0.174 & 0.098 & \underline{0.279} \\

& Random (All) & 0.316 & 0.614 & 0.326 & 0.203 & 0.121 & 0.410 & 0.746 & 0.390 & 0.291 & 0.213 & 0.324 & 0.674 & 0.298 & 0.195 & 0.131 & 0.324 & 0.597 & 0.304 & 0.231 & 0.163 & 0.343 \\

& Random (Ticker) & \underline{0.270} & 0.553 & 0.269 & 0.159 & 0.098 & \underline{0.308} & 0.609 & 0.235 & 0.223 & 0.163 & 0.255 & 0.610 & 0.230 & 0.113 & 0.068 & 0.285 & 0.602 & 0.242 & 0.187 & 0.108 & \underline{0.279} \\

\cline{2-23}

& \textbf{PEV(Mean)} & 0.326 & 0.619 & 0.326 & 0.215 & 0.146 & 0.380 & 0.719 & 0.343 & 0.272 & 0.185 & 0.310 & 0.647 & 0.285 & 0.185 & 0.121 & 0.316 & 0.618 & 0.283 & 0.220 & 0.143 & 0.333 \\

& \textbf{STPEV(Mean)} & \underline{0.270} & 0.584 & 0.245 & 0.152 & 0.099 & 0.310 & 0.640 & 0.270 & 0.201 & 0.129 & \underline{0.243} & 0.592 & 0.219 & 0.104 & 0.057 & 0.278 & 0.599 & 0.236 & 0.183 & 0.095 & \textbf{0.275} \\

\hline\hline

\multirow{10}{*}{2023}  & GPT4o Pred (8 shot) & 
0.34 & 0.9 & 0.217 & 0.146 & 0.097 & 0.307 & 0.775 & 0.225 & 0.133 & 0.097 & 0.324 & 0.613 & 0.325 & 0.211 & 0.147 & 0.301 & 0.627 & 0.259 & 0.177 & 0.143 & 0.318 \\

& Gemini Pred (8 shot) & 0.325 & 0.825 & 0.25 & 0.124 & 0.103 & 0.296 & 0.731 & 0.222 & 0.141 & 0.089 & 0.32 & 0.627 & 0.355 & 0.172 & 0.128 & 0.277 & 0.602 & 0.247 & 0.153 & 0.105 & 0.305 \\

& Vanilla (OpenAI) & 0.268 & 0.663 & 0.197 & 0.109 & 0.104 & 0.274 & 0.643 & 0.235 & 0.134 & 0.084 & 0.226 & 0.439 & 0.249 & 0.129 & 0.088 & 0.258 & 0.557 & 0.222 & 0.142 & 0.111 & 0.257 \\

& Vanilla (Gecko) & 0.257 & 0.659 & 0.189 & 0.104 & 0.076 & 0.266 & 0.619 & 0.226 & 0.132 & 0.089 & 0.215 & 0.408 & 0.213 & 0.126 & 0.113 & \textbf{0.229} & 0.473 & 0.208 & 0.131 & 0.105 & \textbf{0.242} \\

& GPT4o (Summarization) & 0.262 & 0.642 & 0.202 & 0.108 & 0.097 & 0.266 & 0.634 & 0.233 & 0.121 & 0.076 & 0.220 & 0.405 & 0.226 & 0.144 & 0.103 & 0.250 & 0.542 & 0.214 & 0.145 & 0.098 & 0.249 \\

& GPT4o (Task-specific) & 0.249 & 0.634 & 0.188 & 0.097 & 0.078 & 0.262 & 0.621 & 0.221 & 0.125 & 0.080 & 0.220 & 0.418 & 0.209 & 0.142 & 0.110 & 0.253 & 0.542 & 0.220 & 0.140 & 0.110 & 0.246 \\

& Gemini (Summarization) & 0.262 & 0.629 & 0.194 & 0.113 & 0.114 & 0.260 & 0.624 & 0.213 & 0.129 & 0.074 & \textbf{0.210} & 0.415 & 0.224 & 0.119 & 0.083 & \underline{0.238} & 0.523 & 0.210 & 0.127 & 0.092 & 0.243 \\

& Gemini (Task-specific) & 0.262 & 0.637 & 0.204 & 0.115 & 0.092 & 0.269 & 0.622 & 0.228 & 0.136 & 0.088 & \textbf{0.210} & 0.408 & 0.205 & 0.126 & 0.102 & 0.244 & 0.528 & 0.211 & 0.133 & 0.105 & 0.246 \\

& Random (All) & 0.317 & 0.729 & 0.248 & 0.163 & 0.130 & 0.352 & 0.752 & 0.290 & 0.213 & 0.152 & 0.280 & 0.481 & 0.265 & 0.193 & 0.182 & 0.279 & 0.566 & 0.251 & 0.155 & 0.143 & 0.307 \\

& Random (Ticker) & \underline{0.247} & 0.633 & 0.182 & 0.095 & 0.080 & \underline{0.255} & 0.596 & 0.208 & 0.130 & 0.088 & 0.228 & 0.438 & 0.212 & 0.133 & 0.130 & \underline{0.238} & 0.513 & 0.203 & 0.124 & 0.110 & \textbf{0.242} \\

\cline{2-23}

& \textbf{PEV(Mean)} & 0.309 & 0.725 & 0.236 & 0.150 & 0.124 & 0.330 & 0.723 & 0.279 & 0.186 & 0.134 & 0.262 & 0.463 & 0.249 & 0.172 & 0.166 & 0.278 & 0.584 & 0.245 & 0.148 & 0.134 & 0.295 \\

& \textbf{STPEV(Mean)} & \textbf{0.239} & 0.611 & 0.180 & 0.093 & 0.074 & \textbf{0.253} & 0.601 & 0.209 & 0.122 & 0.081 & 0.227 & 0.432 & 0.215 & 0.132 & 0.130 & 0.246 & 0.520 & 0.215 & 0.133 & 0.118 & \textbf{0.242} \\

\hline

\end{tabular}
}
\caption{The overall performance on DEC.}

\label{table: The Performance for DEC on Text}
\end{table*}

\subsection{Evaluations on DEC Dataset}  
\label{Evaluations on DEC Dataset}

\paragraph{Evaluation settings.} Following Section~\ref{Evaluations on EC and MAEC Datasets}, we evaluate various transcript-based models on DEC. These include LLM few-shot direct prediction, using 8 prior \textit{(earnings call transcripts, volatility)} pairs, and LLM embeddings, comprising two vanilla text embeddings\footnote{Voyage has been shown to perform poorly in Table~\ref{table:The Overall Performance}, we report Vanilla(Voyage) results on DEC in Appendix~\ref{Appendix Results for Transcripts-based Models.}.} and four fine-grained text embeddings. Additionally, PEV(Mean) and STPEV(Mean) are applied on DEC.

To explore the role of semantics in transcripts and how ticker identity is purely reflected in transcript-level\footnote{The main difference between PEV(Mean) and STPEV(Mean) lies in the ticker identity, which we aim to embed solely at the "transcript-level" representation.}, we introduce two random embeddings for comparison with transcript embeddings:
\begin{itemize}[topsep=3pt, partopsep=0pt, itemsep=0pt, parsep=0pt]
    \item \textbf{Random(All):} Each transcript is assigned a random embedding. This approach effectively removes both semantics and ticker identity.
    \item \textbf{Random(Ticker):} Transcripts belonging to the same ticker are assigned the same randomly generated embedding. This removes semantics while preserving ticker identity.
\end{itemize}

%

We only report the results from 2021 to 2023 in Table~\ref{table: The Performance for DEC on Text}, as we suffer from sparse overlapping earnings for the years 2019 and 2020.  The full results can be found in Appendix~\ref{Appendix Results for Transcripts-based Models.}.


\paragraph{\textit{Performance of different models.}}
As shown in Table~\ref{table: The Performance for DEC on Text}, direct prediction by GPT-4o and Gemini yield the worst performance with the MSEs of 0.345 and 0.339. Transcripts-based models exhibit comparable performance among themselves: the best model, Gemini(Summarization), and the worst model, Vanilla(Gecko), achieve average MSEs of 0.260 and 0.269, respectively. Further analysis in Appendix~\ref{Appendix Clustering among Different Text} shows that different LLMs fail to generate distinct texts for the same transcripts, whereas two strategies, \textit{Summarization} and \textit{task-specific}, can differentiate transcripts effectively.

Despite containing no semantics or insights in their input, the Random(Ticker) embeddings and STPEV(Mean), with overall MSEs of 0.265 and 0.252 respectively, with STPEV(Mean) surpassing various semantically meaningful transcripts models. Moreover, models that lack ticker identity, such as Random(All) embeddings and the PEV(Mean), with average MSEs of 0.323 and 0.307 respectively, consistently underperform compared to other models that include ticker identity across all quarters. 

These findings suggest that \textit{the ticker identity plays a determinant role}, while \textit{the semantic content of transcripts is plausibly not useful}. This motivates us to raise the following hypothesis:
\begin{center}
     \textbf{Are the semantics of transcripts converging toward ticker identity?} 
\end{center}


\paragraph{\textit{Representation-level comparisons between within-ticker group and all-dataset group.}}
\begin{table}[ht]
\centering
\resizebox{\columnwidth}{!}{%
\begin{tabular}{|cc|cc|}
\hline
\multirow{2}{*}{Ticker Identity} & 
\multirow{2}{*}{Model} & 
\multicolumn{2}{c|}{Cosine Similarity} \\
& & Within-Ticker & All-dataset \\
\hline

\multirow{8}{*}{With} & Vanilla (OpenAI) & 0.9 & 0.7\\
& Vanilla (Gecko) & 0.958 & 0.865\\
& GPT4o (Summarization) & 0.92 & 0.685  \\
& GPT4o (Task-specific) & 0.929 & 0.724\\
& Gemini (Summarization) & 0.931 & 0.713 \\
& Gemini (Task-specific) & 0.918 & 0.728 \\
& Random (Ticker) & 1.0 & 0.752 \\

& \textbf{Average} & \textbf{0.937} & \textbf{0.738} \\
\hline
Without & Random (All) & 0.765 & 0.753\\
\hline
\end{tabular}
}
\caption{The mean cosine similarity between the within-ticker group and the all-dataset group.}
\label{table: The cosine similarity within-ticker V.S. all-dataset}
\end{table}

We compare the cosine similarity (for each earnings record) within individual tickers and across the entire dataset for text embeddings. As shown in Table~\ref{table: The cosine similarity within-ticker V.S. all-dataset}, the within-ticker similarity is consistently higher than the overall similarity when ticker identity is present, even for texts scrutinized by LLMs. This observation aligns with the findings of \citet{liu-etal-2024-beyond}: \textit{the semantics appear largely similar on the surface for financial statements of the same company but correspond to different periods}. Additionally, the similarity across the entire dataset also remains high due to shared structures and characteristics among all earnings. For more details, please refer to Appendix~\ref{Appendix Similarity Comparison}.

This finding is intuitive, as earnings calls for each company tend to follow similar and structured patterns over time. This finding supports the hypothesis that \textit{transcripts primarily capture the ticker identity}, owing to the inherently similar nature of transcripts for the same ticker.

\paragraph{\textit{Prediction-level comparisons between transcripts-based models and STPEV(Mean).}}
In the STPEV(Mean), we use the mean function as an approximate proxy for the prior post-earnings volatility distribution. We thus compare the predictions of different transcripts-based models with those of the STPEV(Mean) and calculate the Pearson correlation coefficients between them. As shown in Table~\ref{table: The pearson correlation coefficient}, The predictions of transcripts-based models and STPEV(Mean) are highly linearly correlated for all models that contain the ticker identity, with an average correlation coefficient of {0.847} over 3 years. This also validates our hypothesis that \textit{transcripts approximate the prior same-ticker post-earnings volatility}, where different transcripts-based texts capture the ticker identity and the mean value is used to represent the prior distribution. Further details are provided in Appendix~\ref{Appendix Predictions Correlation}.

\begin{table}[ht]
\centering
\resizebox{\columnwidth}{!}{%
\begin{tabular}{|cc|ccc|}
\hline
\multirow{2}{*}{Ticker Identity} & 
\multirow{2}{*}{Model} & 
\multicolumn{3}{c|}{Yearly Average} \\
& & 2021 & 2022 & 2023 \\
\hline
\multirow{7}{*}{With} & Vanilla (OpenAI) & 0.87 & 0.873 & 0.866 \\
& Vanilla (Gecko) & 0.753 & 0.825 & 0.81 \\
& GPT4o (Summarization) & 0.799 & 0.792 & 0.848 \\
& GPT4o (Task-specific) & 0.831 & 0.855 & 0.916 \\
& Gemini (Summarization) & 0.852 & 0.854 & 0.867  \\
& Gemini (Task-specific) & 0.816 & 0.843 & 0.867 \\
& Random (Ticker) & 0.786 & 0.925 & 0.946 \\

& \textbf{Average} & \textbf{0.815} & \textbf{0.852} & \textbf{0.874} \\
\hline
Without & Random (All) & 0.183 & 0.007 & 0.004 \\
\hline
\end{tabular}
}
\caption{Pearson correlation coefficients between the predictions of transcripts models and of STPEV(Mean).}
\label{table: The pearson correlation coefficient}
\end{table}

 
 


\section{Conclusion}
In this work, we introduce a dense earnings call dataset: DEC, in which each ticker is represented by 20 earnings. Motivated by the significant drifts of post-earnings volatility and the ticker-specific volatility regime, we propose two training-free baselines, PEV and STPEV. These surpass various transcript-based models through simple mean implementation. We further confirm that transcripts primarily capture ticker identity and approximate the past post-earnings volatility, by representation-level and prediction-level comparisons.

\section*{Limitations}
\paragraph{Limited Earnings Call Representations.} In this work, we rely solely on earnings call transcripts to represent earnings, whereas other formats, such as investor presentations, financial statements, and video or audio webcasts, could also be utilized.
\paragraph{Limited Earnings Call Transcript Representations.} This work primarily utilizes LLM embedding models and fine-grained texts to represent earnings call transcripts alone. However, integrating additional financially meaningful information, such as corporate backgrounds, recent news releases, and analysts’ expectations, could further enhance the richness of transcript representations.

\section*{Acknowledgements} We thank EOD\footnote{\url{https://eodhd.com/}} for providing reliable earnings information and price records. We also thank Seeking Alpha\footnote{\url{https://seekingalpha.com/}} for supplying sector-level information and earnings call transcripts.



\bibliography{anthology,custom}
\bibliographystyle{acl_natbib}

\appendix
\clearpage

\section{Dataset Details}

\subsection{Missing BeforeAfterMarket: An Example}
\label{Appendix Important BeforeAfterMarket}
Previous studies have overlooked the fact that earnings can be released \textit{before the market opens}, which is 9 AM in U.S. exchanges. Here is a specific example in EC dataset: Target (TGT) Q3 2017 Earnings Call\footnote{\url{https://seekingalpha.com/article/4125212-target-tgt-q3-2017-results-earnings-\\call-transcript}}. This earnings release occurred at 8:00 AM ET on Nov. 15, 2017. Since market participants were provided with this earnings disclosures and traded based on it during trading hours on Nov. 15, the first post earnings day should be considered as \textit{Nov. 15} rather than \textit{Nov. 16}. According to the definition of post-earnings volatility, the three-day volatility should account for the trading days \textit{\{Nov. 15, Nov. 16, and Nov. 17\}}, not \textit{\{Nov. 16, Nov. 17, and Nov. 20\}}. Consequently, the volatility should be recalculated as -2.726, whereas previous studies incorrectly recorded it as -3.703. 

Regrettably, both the EC and MAEC datasets exhibit critical errors when calculating volatility for earnings released before the market opening. Specifically, the EC, MAEC15, and MAEC16 datasets contain 368, 395, and 584 earnings released before market opening, accounting for 69.3\%, 64.3\%, and 64.2\% of the datasets, respectively. Thus, we believe that the volatility values used in prior studies are unreliable.

\subsection{DEC Dataset Details}
\label{Appendix Dataset DEC}

To ensure a diverse representation of tickers in the U.S. markets, we selected 11 sectors: \textit{Technology, Healthcare, Industrial, Utility, Real Estate, Basic Materials, Financial Services, Consumer Discretionary, Consumer Staples, Communication Services, and Energy}. Each sector exhibits distinct characteristics in response to earnings calls, driven by differences in business models, investor expectations, and macroeconomic influences.

For each sector, we selected companies that are among the top 10 holdings in sector-specific ETFs. For example, in the \textit{technology} sector, the top 10 companies held by the \textbf{XLK ETF}\footnote{\url{https://seekingalpha.com/symbol/XLK}} include \textit{Apple Inc., NVIDIA Corp., Microsoft Corp., Broadcom Inc., Salesforce Inc., Oracle Corp., Cisco Systems Inc., Adobe Inc., Accenture PLC Class A, and Advanced Micro Devices Inc}. In this way, we further ensure that DEC includes most of the representative companies in the U.S. while maintaining diversity.

After identifying 110 tickers, we merged earnings call transcripts\footnote{\url{https://seekingalpha.com/}} with price records\footnote{\url{https://eodhd.com/}}. During this process, some tickers were excluded due to various reasons, such as incomplete earnings cycles (fewer than 20 earnings records) or missing price data or the \texttt{beforeAfterMarket} attribute.

\begin{figure}[!htb]  
    \centering
    \begin{subfigure}{.48\textwidth}
        \centering
        \includegraphics[width=\linewidth]{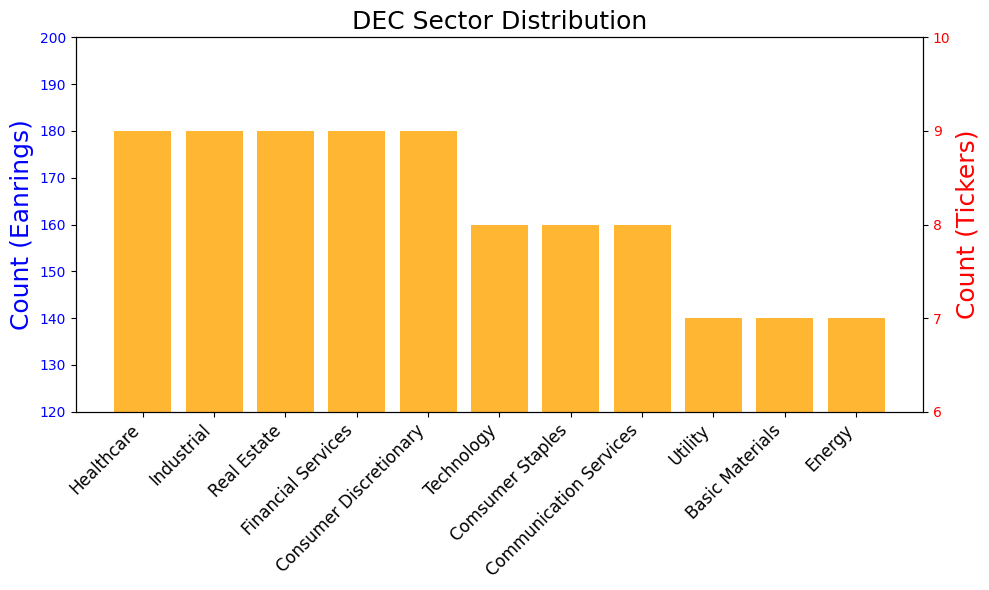}
    \end{subfigure}%
    \hfill 
    \caption{DEC Sector Distribution}
\label{Figure: DEC Sector Distribution}
\end{figure}

Ultimately, we retained 90 tickers across 11 sectors, with each ticker containing 20 earnings records spanning from the first quarter of 2019 to the last quarter of 2023, resulting in a total of 1,800 earnings records. The sector distribution is illustrated in Figure~\ref{Figure: DEC Sector Distribution}.

\begin{figure*}[!htb]  
    \centering

     \begin{subfigure}{.24\textwidth}
        \centering
        \includegraphics[width=\linewidth]{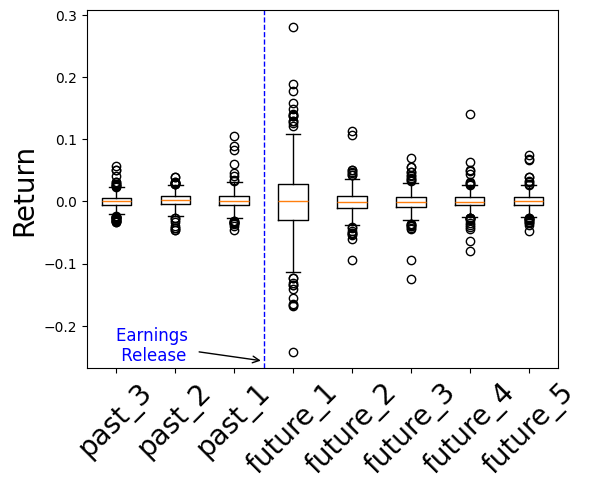}
        \caption{EC}
        \label{EC Return}
    \end{subfigure}%
    \hfill 
    \begin{subfigure}{.24\textwidth}
        \centering
        \includegraphics[width=\linewidth]{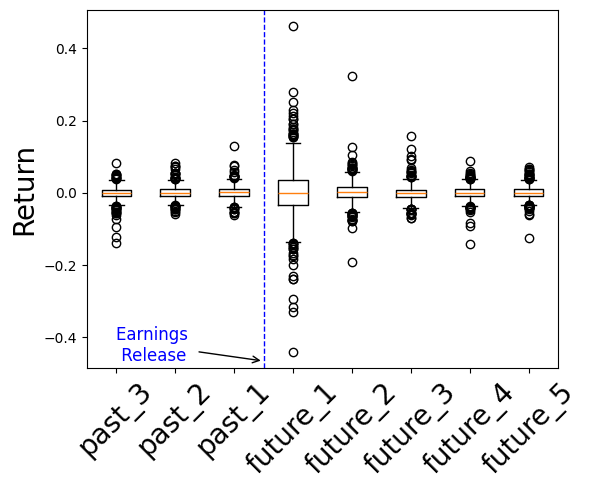}
        \caption{MAEC-15}
        \label{MAEC-15 Return}
    \end{subfigure}%
    \hfill 
    \begin{subfigure}{.24\textwidth}
        \centering
        \includegraphics[width=\linewidth]{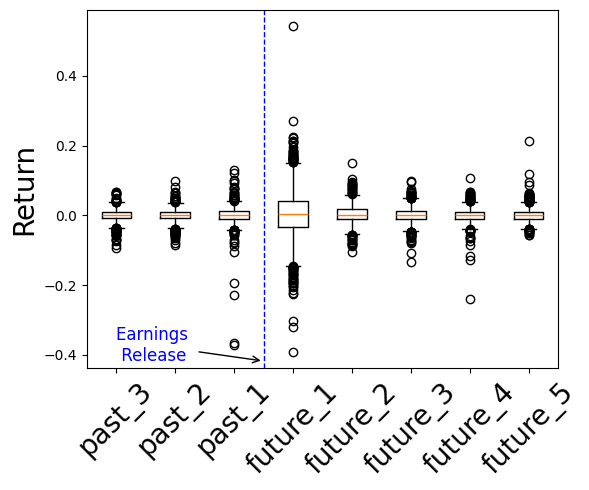}
        \caption{MAEC-16}
        \label{MAEC16 Return}
    \end{subfigure}
    \begin{subfigure}{.24\textwidth}
        \centering
        \includegraphics[width=\linewidth]{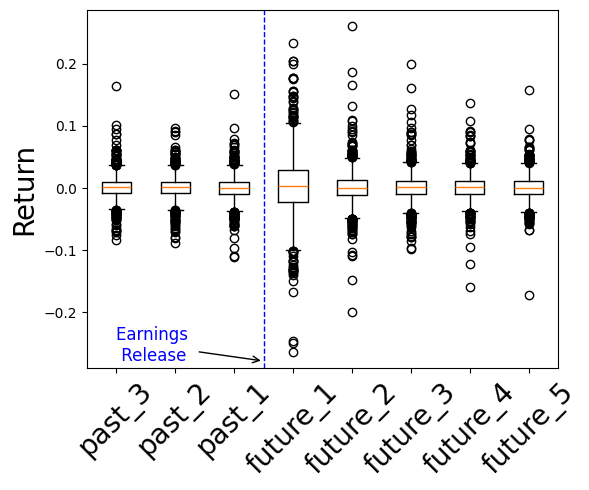}
        \caption{DEC}
        \label{DEC Return}
    \end{subfigure}

   \caption{Comparison of \textit{returns} before and after earnings announcements. Earnings are released between the day labeled \texttt{past\_1} and the day labeled \texttt{future\_1}. The absolute return on \texttt{future\_1} is significantly higher than on pre-earnings days or other subsequent post-earnings days.}
\label{Figure: Return Comparison Before V.S. After E}
\end{figure*}

\begin{figure*}[!htb]  
    \centering

     \begin{subfigure}{.32\textwidth}
        \centering
        \includegraphics[width=\linewidth]{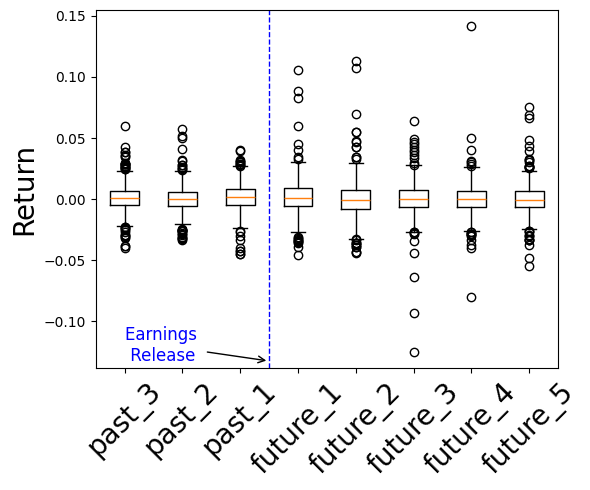}
        \caption{EC}
        \label{w/o EC Return}
    \end{subfigure}%
    \hfill 
    \begin{subfigure}{.32\textwidth}
        \centering
        \includegraphics[width=\linewidth]{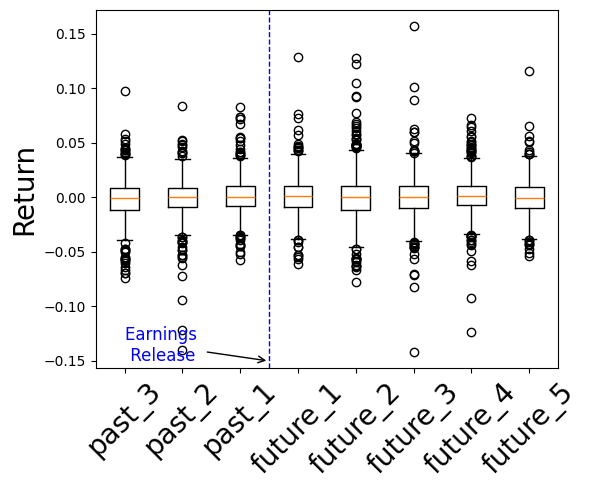}
        \caption{MAEC-15}
        \label{w/o MAEC-15 Return}
    \end{subfigure}%
    \hfill 
    \begin{subfigure}{.32\textwidth}
        \centering
        \includegraphics[width=\linewidth]{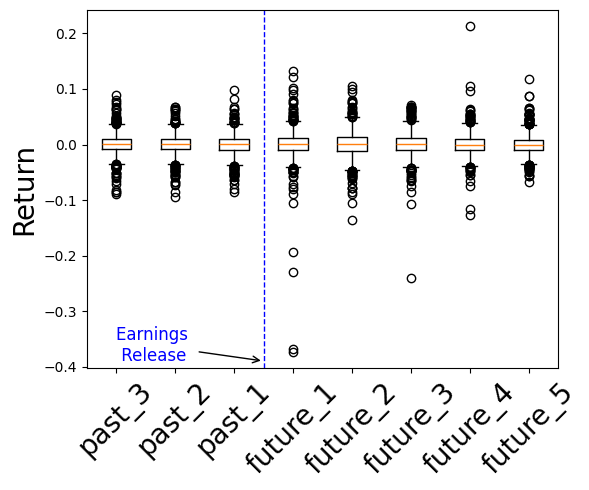}
        \caption{MAEC-16}
        \label{w/o MAEC16 Return}
    \end{subfigure}

   \caption{Comparison of  \textit{returns (without beforeAfterMarket adjustment)} before and after earnings announcements. Earnings are released between the day labeled \texttt{past\_1} and the day labeled \texttt{future\_1}. The increased absolute \texttt{future\_1} tends to diminish or disappear, compared to Figure~\ref{Figure: Return Comparison Before V.S. After E}.}
   
\label{Figure: Original Return Comparison Before V.S. After E}
\end{figure*}

\begin{figure*}[!htb]  
    \centering
    
    \begin{subfigure}{.33\textwidth}
        \centering
        \includegraphics[width=\linewidth]{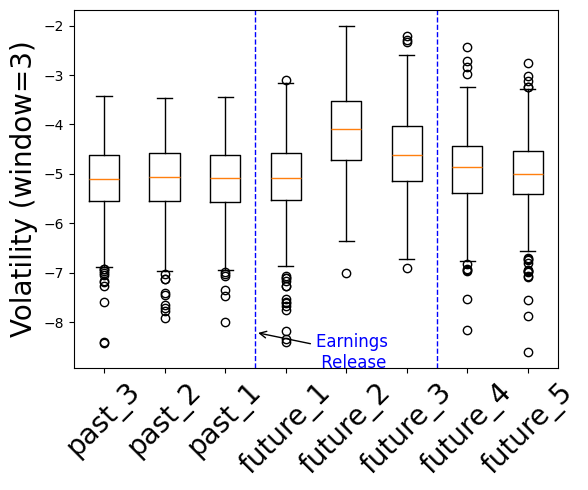}
        \caption{EC}
        \label{Original EC E VS NE}
    \end{subfigure}%
    \hfill 
    \begin{subfigure}{.33\textwidth}
        \centering
        \includegraphics[width=\linewidth]{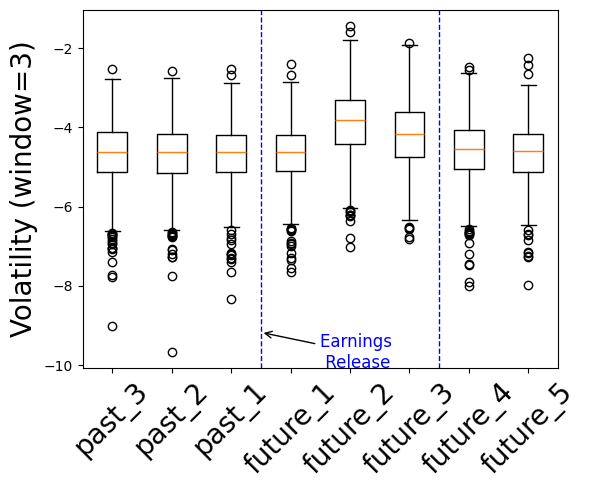}
        \caption{MAEC-15}
        \label{Original MAEC-15 E VS NE}
    \end{subfigure}%
    \hfill 
    \begin{subfigure}{.33\textwidth}
        \centering
        \includegraphics[width=\linewidth]{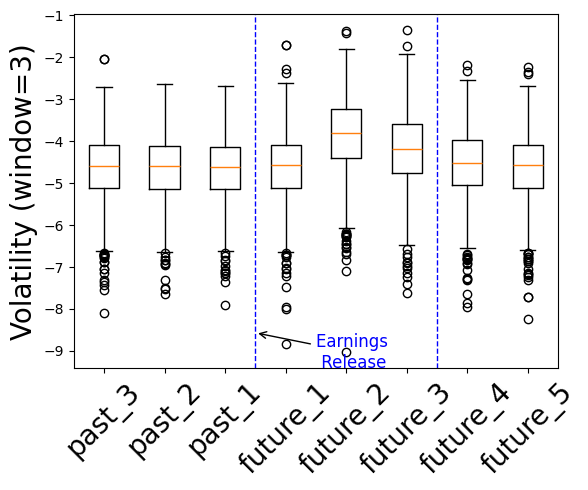}
        \caption{MAEC-16}
        \label{Original MAEC-16 E V.S. NE}
    \end{subfigure}
    \caption{Comparison of three-day volatility \textit{(without beforeAfterMarket adjustment)} before and after earnings announcements. Earnings are released between the day labeled \texttt{past\_1} and the day labeled \texttt{future\_1}. Figure~\ref{Figure: Volatility Comparison Before V.S. After E} displays the same comparison except  \textit{with beforeAfterMarket adjustment}.}
\label{Figure: Original Volatility Comparison Before V.S. After E}
\end{figure*}

\section{Two Observations from DEC}
\subsection{Post Earnings Volatility Distribution Drift}
\label{Appendix Post Earnings Volatility Distribution Drift}

Figure~\ref{Figure: Return Comparison Before V.S. After E} compares the daily returns before and after earnings for the EC, MAEC, and DEC datasets. The absolute return on the first day after earnings, \( r_{future\_1} \), is significantly higher than on other days. In contrast, Figure~\ref{Figure: Original Return Comparison Before V.S. After E} compares the daily returns \textit{without beforeAfterMarket adjustment} from the original EC and MAEC datasets\footnote{\url{https://github.com/hankniu01/KeFVP/tree/main/price_data}}. In this case, the effect tends to diminish or disappear due to the incorrect time used in identifying \( r_{future\_1} \). This observation further validates the importance of the \texttt{beforeAfterMarket} attribute. 

In Section~\ref{Distribution Shift After Earnings}, we conclude that days involving the volatility calculation of \( r_{future\_1} \) (the first day after earnings) exhibit significantly higher volatility compared to other days. Figure~\ref{Figure: Volatility Comparison Before V.S. After E} illustrates this pattern for a 3-day window. In contrast, Figure~\ref{Figure: Original Volatility Comparison Before V.S. After E} compares the same trend \textit{without the beforeAfterMarket adjustment} using the original EC and MAEC datasets, where the increased volatility within the 3-day window is observed to dilute. This further highlights the importance of incorporating the \texttt{beforeAfterMarket} attribute.

\begin{figure*}[!htb]  
    \centering
    
    \begin{subfigure}{.24\textwidth}
        \centering
        \includegraphics[width=\linewidth]{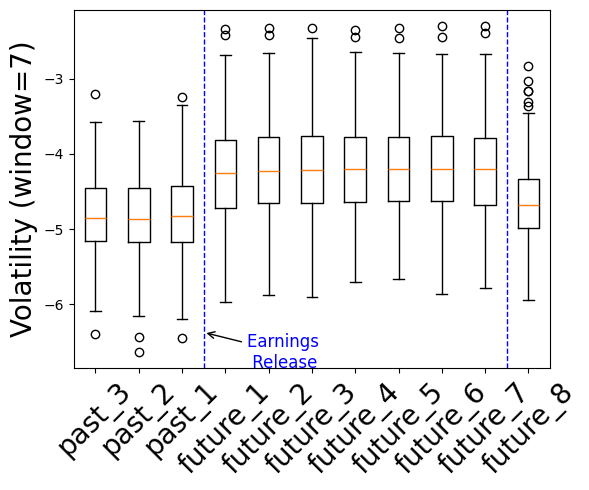}
        \caption{EC}
        \label{EC E V.S. NE win7}
    \end{subfigure}%
    \hfill 
    \begin{subfigure}{.24\textwidth}
        \centering
        \includegraphics[width=\linewidth]{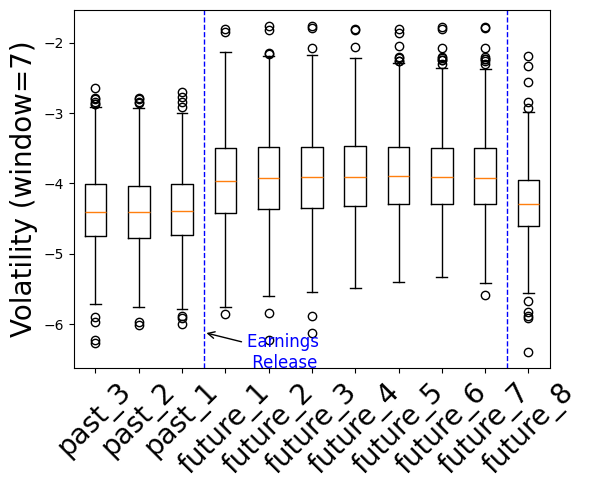}
        \caption{MAEC-15}
        \label{MAEC-15 E V.S. NE win7}
    \end{subfigure}%
    \hfill 
    \begin{subfigure}{.24\textwidth}
        \centering
        \includegraphics[width=\linewidth]{plots/E_VS_NE/MAEC15_Volatility_E_VS_NE_win7.png}
        \caption{MAEC-16}
        \label{MAEC-16 E V.S. NE win7}
    \end{subfigure}
    \begin{subfigure}{.24\textwidth}
        \centering
        \includegraphics[width=\linewidth]{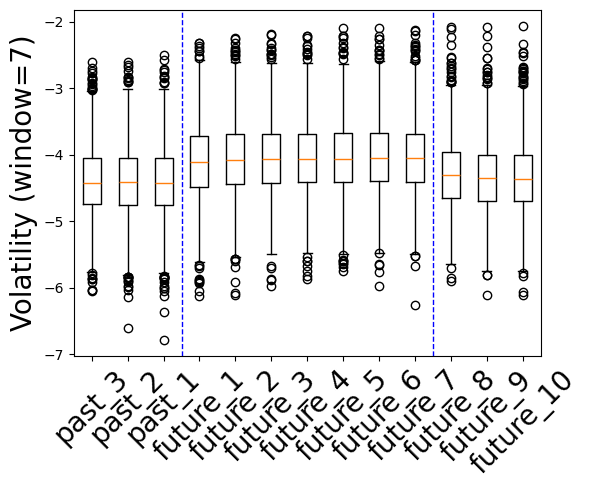}
        \caption{DEC}
        \label{DEC E V.S. NE win7}
    \end{subfigure}
    \caption{Comparison of seven-day volatility before and after earnings announcements. Earnings are released between the day labeled \texttt{past\_1} and the day labeled \texttt{future\_1}. Days where the volatility calculation involves the return of \texttt{future\_1} exhibit significantly higher volatility compared to others.}
\label{Figure: Volatility Comparison Before V.S. After E win7}
\end{figure*}

\begin{figure*}[!htb]  
    \centering
    
    \begin{subfigure}{.33\textwidth}
        \centering
        \includegraphics[width=\linewidth]{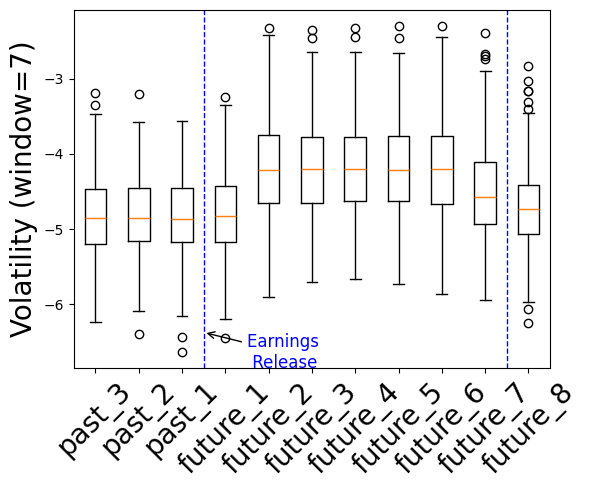}
        \caption{EC}
        \label{Original EC E VS NE win7}
    \end{subfigure}%
    \hfill 
    \begin{subfigure}{.33\textwidth}
        \centering
        \includegraphics[width=\linewidth]{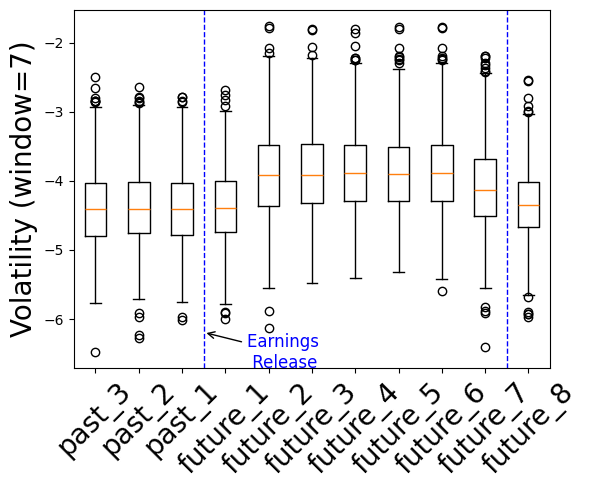}
        \caption{MAEC-15}
        \label{Original MAEC-15 E VS NE win7}
    \end{subfigure}%
    \hfill 
    \begin{subfigure}{.33\textwidth}
        \centering
        \includegraphics[width=\linewidth]{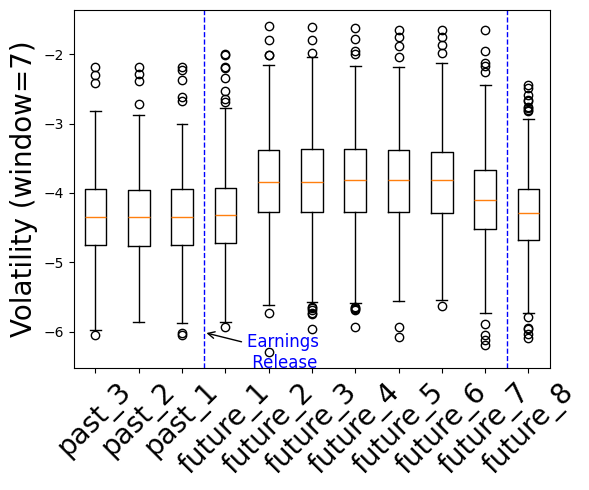}
        \caption{MAEC-16}
        \label{Original MAEC-16 E V.S. NE win7}
    \end{subfigure}
    \caption{Comparison of seven-day volatility \textit{(without beforeAfterMarket adjustment)} before and after earnings announcements. Earnings are released between the day labeled \texttt{past\_1} and the day labeled \texttt{future\_1}. Figure~\ref{Figure: Volatility Comparison Before V.S. After E win7} displays the same comparison except  \textit{with beforeAfterMarket adjustment}.}
\label{Figure: Original Volatility Comparison Before V.S. After E win7}
\end{figure*}

Furthermore, Figure~\ref{Figure: Volatility Comparison Before V.S. After E win7} demonstrates that the post-earnings volatility distribution drift persists for a 7-day window. In contrast, Figure~\ref{Figure: Original Volatility Comparison Before V.S. After E win7} presents the same comparison \textit{without the beforeAfterMarket adjustment}, where the phenomenon diminishes at the beginning and end of the volatility window. This further underscores the importance of incorporating the \texttt{beforeAfterMarket} attribute.

We do not present volatility comparisons for window sizes of 15 and 30, as they could not fit into a single figure due to space constraints. Nevertheless, the same distribution drift is observed for these window sizes due to the increased absolute values of \( r_{future\_1} \).


\begin{figure*}[!htb]  
    \centering
    \begin{subfigure}{.24\textwidth}
        \centering
    \includegraphics[width=\linewidth]{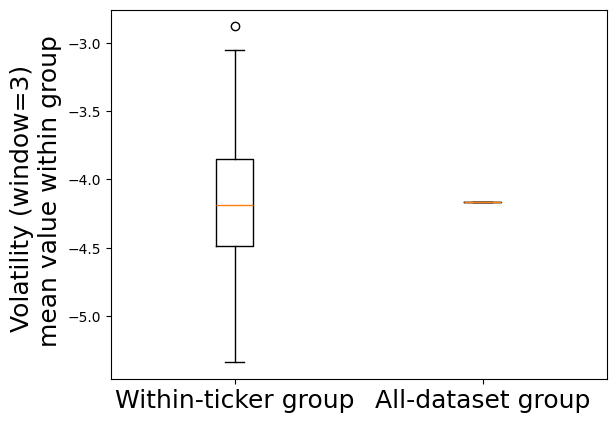}
        \caption{EC}
        \label{fig:EC T VS O}
    \end{subfigure}%
    \hfill 
    \begin{subfigure}{.24\textwidth}
        \centering
        \includegraphics[width=\linewidth]{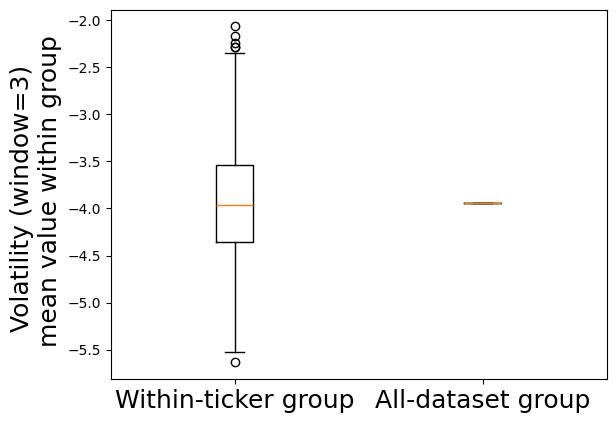}
        \caption{MAEC-15}
        \label{fig:MAEC-15 T VS O}
    \end{subfigure}%
    \hfill 
    \begin{subfigure}{.24\textwidth}
        \centering
        \includegraphics[width=\linewidth]{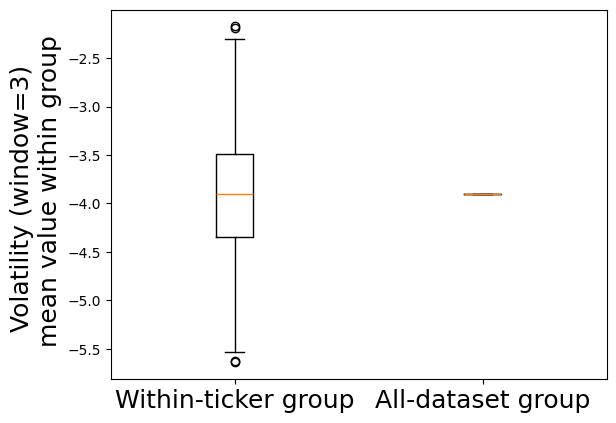}
        \caption{MAEC-16}
        \label{fig:MAEC-16 T VS O}
    \end{subfigure}
    \begin{subfigure}
    {.24\textwidth}
        \centering
        \includegraphics[width=\linewidth]{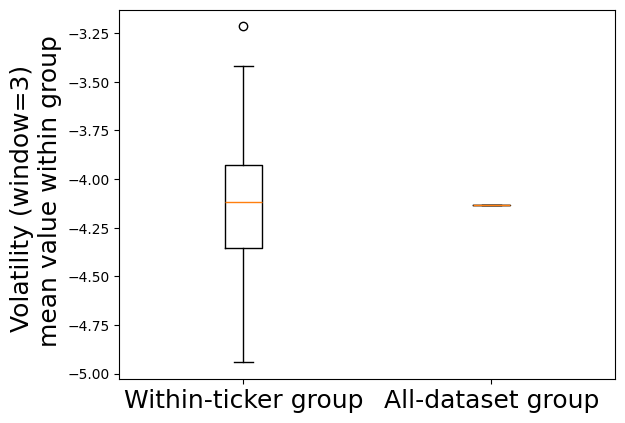}
        \caption{DEC}
        \label{fig:DEC T VS O}
    \end{subfigure}
    \caption{Comparison of the mean of three-day volatility between the within-ticker group and the all-dataset group.}
    \label{fig:Mean Comparison T V.S. O}
\end{figure*}

\subsection{Ticker-Specific Volatility Regime}
\label{Appendix Ticker-Specific Volatility Regime}
The ticker-specific volatility regime posits that each company tends to follow a distinct post-earning volatility distribution. We term this phenomenon as \textit{Volatility Signature}, which likely arises from intrinsic company characteristics that remain relatively stable over short periods. These characteristics may include industry and sector classification, operational dynamics, company size and market position, and financial structure. Such stable properties act as anchors, mitigating extreme volatility fluctuations and maintaining predictable patterns of post-earnings volatility, even in response to periodic financial disclosures. 

To further illustrate the volatility signature, Figure~\ref{fig:Mean Comparison T V.S. O} compares the mean values of three-day volatility between the within-ticker group and the all-dataset group. This analysis is particularly relevant, as the mean function is primarily used and benchmarked against baselines in Section~\ref{Main Results}. The figure reveals significant variation in mean values across tickers (as shown on the y-axis), underscoring the motivation for introducing STPEV as an enhancement to PEV.

\section{Augmentation on EC and MAEC}
\label{Appendix Augmentation on EC and MAEC}

\begin{table}[ht]
\centering
\resizebox{\columnwidth}{!}{%
\begin{tabular}{c|cc|cc|cc|}
\hline
Dataset & \multicolumn{2}{c|}{EC} & \multicolumn{2}{c}{MAEC-15} & \multicolumn{2|}{c}{MAEC-16} \\
\hline
Type & Original & Augmented & Original & Augmented  & Original & Augmented  \\
\hline
Range & 2017-2017 & 2012-2017 & 2015-2015 & 2010-2015 & 2015-2016 & 2011-2016  \\
\hline
Count (Train) & 179 & 2195 & 94 & 3192 & 215 & 5765 \\ 
Count (Test) & 112 & 112 & 154 & 154 & 280 & 280 \\
OET & 1.598 & 19.775 & 0.61 & 20.727 & 0.768 & 20.812 \\
\hline
\end{tabular}%
}
\caption{EC and MAEC statistics between the original and the augmented for STPEV. OET is defined as the proportion of overlapping training earnings over testing tickers defined in equation~\ref{eq:rot_formula}.}
\label{table: The Vanilla V.S. Augmented Datasets Comparison}
\end{table}

Since the PEV and STPEV take historical post-earnings volatility as input, the current EC and MAEC datasets, which lack sufficient previous same-ticker earnings records, must be left-extended to earlier years. Table~\ref{table: The Vanilla V.S. Augmented Datasets Comparison} compares the data statistics of testing tickers overlapped training earnings relative to testing tickers (OET) before and after augmentation. It is evident that the OET values are significantly improved by left-extension.


\section{Evaluations and Analysis on DEC}
\label{Appendix Evaluations and Analysis on DEC}

\subsection{Implementation Details for Transcripts-based Models}
One NVIDIA L40 GPU is used for the transcript-based models. The learning rate is set to 1e-4, with a batch size of 32 and a random seed of 2021. The models are trained for up to 10 epochs using early-stopping techniques. All results are based on a single run.

The embedding dimensions for OpenAI, Gecko, and Voyage embeddings models are 3071, 768, and 1024, respectively. The 2-layer MLP has a hidden size of 512 in the middle layer.

When evaluating a quarter on DEC, all previous quarters are randomly split into training and validation sets with a 2:1 ratio, using a random state of 42. The preliminary experiments demonstrate that this approach outperforms the following settings:
\begin{itemize}
    \item Using all previous quarters, with the first two-thirds as the training set and the last one-third as the validation set.
    \item Using the previous \textit{three} quarters only, randomly split into training and validation sets with a 2:1 ratio and random state of 42.
    \item Using the previous \textit{three} quarters, with the first two-thirds as the training set and the last one-third as the validation set.
\end{itemize}

\subsection{Transcripts-based Models with LLM}
\label{Appendix LLM Fine-grained Texts}

\paragraph{LLM Direct Prediction}
We utilize few-shot learning to prompt LLMs for direct volatility prediction, providing the task description and prior \textit{(earnings call transcripts, volatility)} pairs within the prompt. For EC and MAEC, three randomly selected pairs are used as demonstrations, while for DEC, eight ticker-specific prior pairs are included as demonstrations. The prompt template is illustrated in Figure~\ref{fig: Prompt template for LLM direct volatility prediction}.

\begin{figure}[t]
\begin{tcolorbox}[colback=orange!5!white,colframe=orange!75!black,title= Prompt for LLMs Direct Volatility Prediction, width=\columnwidth, fontupper=\scriptsize]

Company \textit{{ticker}} has just released its earnings transcript.
Our primary goal is to predict the volatility for the next \textit{\{prediction\_window\}} trading days.
\\

Let me first clarify our target:
volatility = log(std(r1, r2, ..., rn)), where ri is the return on day i in the future. \\
In general, higher volatility means more dramatic price fluctuations, indicating a more volatile market. \\

To help you understand the task, here are some previous examples of earnings call transcripts
and their corresponding volatility values for \textit{{ticker}} over the past 2 years (a total of 8 earnings).
The most recent pair is labeled as previous 1, representing the latest past earnings, while previous 8 refers to the oldest past earnings. \\

Previous 1 (transcripts, volatility) pair for 	\textit{{ticker}}:
- Transcript (start): {}
- Transcript (end).
- Target (volatility for the next  	ext{{prediction\_window}} trading days):  	\textit{{volatility}} \\

Previous 2 (transcripts, volatility) pair for 	\textit{{ticker}}:
- Transcript (start): {}
- Transcript (end).
- Target (volatility for the next  	ext{{prediction\_window}} trading days):  	\textit{{volatility}} \\
... \\
... \\
... \\
Previous 8 (transcripts, volatility) pair for 	\textit{{ticker}}:
- Transcript (start): {}
- Transcript (end).
- Target (volatility for the next  	ext{{prediction\_window}} trading days):  	\textit{{volatility}} \\

Now that you've reviewed the goal and examples, here's the current earnings call transcript for analysis: \\
- Transcript (start): \textit{{current\_transcripts}}
- Transcript (end). \\

Let's proceed step by step:
1. Recognize patterns for using the transcripts of \textit{{ticker}} to predict volatility. \\
2. Perform a comparative analysis of the current earnings transcript with the previous examples, as quarter-to-quarter performance is critical for earnings. \\
3. Use your identified patterns and comparative analysis to predict the volatility for the next \textit{{prediction\_window}} days. \\

Details about your reasoning process are highly appreciated.

\end{tcolorbox}
\caption{Prompt for LLMs direct volatility prediction.}
\label{fig: Prompt template for LLM direct volatility prediction}
\end{figure}

\begin{figure}[t]
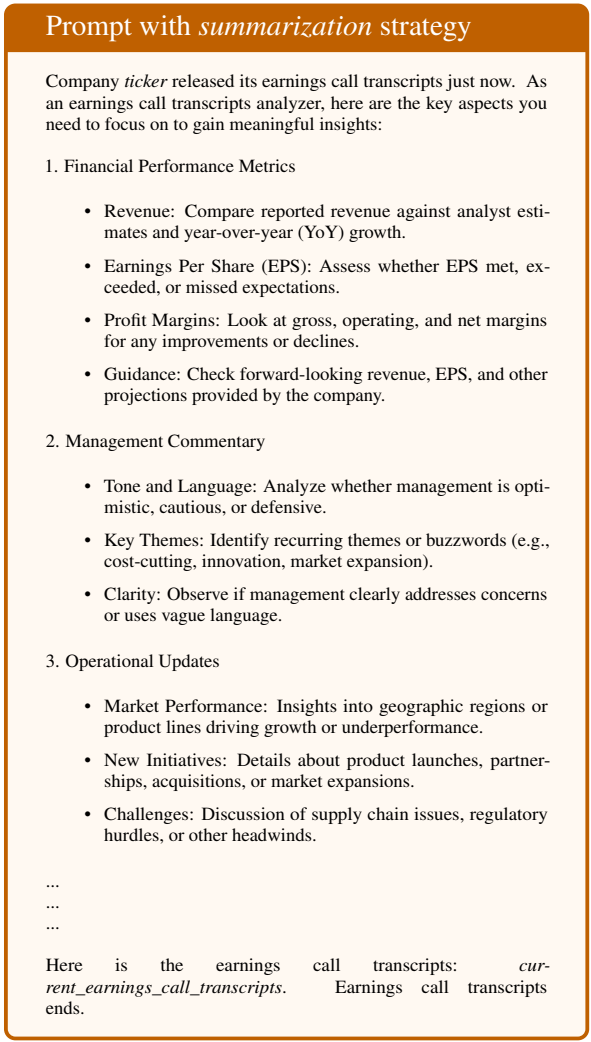

\begin{tcolorbox}[colback=orange!5!white,colframe=orange!75!black,title= Prompt with \textit{summarization} strategy ,width=\columnwidth, fontupper=\scriptsize]

Company \textit{ticker} released its earnings call transcripts just now. \
As an earnings call transcripts analyzer, here are the key aspects you need to focus on to gain meaningful insights: \\

1. Financial Performance Metrics
    \begin{itemize}[itemsep=0pt,]
        \item Revenue: Compare reported revenue against analyst estimates and year-over-year (YoY) growth.
        \item Earnings Per Share (EPS): Assess whether EPS met, exceeded, or missed expectations.
        \item Profit Margins: Look at gross, operating, and net margins for any improvements or declines.
        \item Guidance: Check forward-looking revenue, EPS, and other projections provided by the company.
    \end{itemize}

2. Management Commentary
    \begin{itemize}[itemsep=0pt]
        \item Tone and Language: Analyze whether management is optimistic, cautious, or defensive.
        \item Key Themes: Identify recurring themes or buzzwords (e.g., cost-cutting, innovation, market expansion).
        \item Clarity: Observe if management clearly addresses concerns or uses vague language.
    \end{itemize}

3. Operational Updates
    \begin{itemize}[itemsep=0pt]
        \item Market Performance: Insights into geographic regions or product lines driving growth or underperformance.
        \item New Initiatives: Details about product launches, partnerships, acquisitions, or market expansions.
        \item Challenges: Discussion of supply chain issues, regulatory hurdles, or other headwinds.
    \end{itemize}

... \\
... \\
... \\

Here is the earnings call transcripts: \textit{current\_earnings\_call\_transcripts}. 
Earnings call transcripts ends. 

\end{tcolorbox}
\caption{Prompt with \textit{summarization} strategy.}
\label{fig: Prompt for LLMs with summarization strategy.}
\end{figure}

\begin{figure}[t]
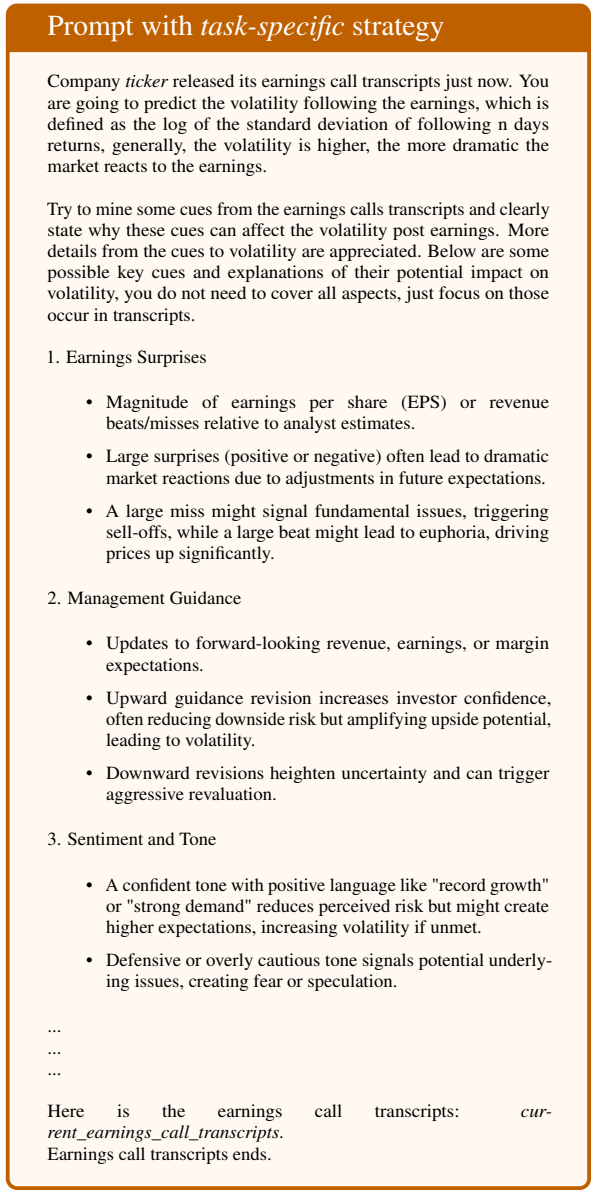

\begin{tcolorbox}[colback=orange!5!white,colframe=orange!75!black,title= Prompt with \textit{task-specific} strategy ,width=\columnwidth, fontupper=\scriptsize]
Company \textit{ticker} released its earnings call transcripts just now. You are going to predict the volatility following the earnings, which is defined as the log of the standard deviation of following n days returns, generally, the volatility is higher, the more dramatic the market reacts to the earnings. \\

Try to mine some cues from the earnings calls transcripts and clearly state why these cues can affect the volatility post earnings. More details from the cues to volatility are appreciated. Below are some possible key cues and explanations of their potential impact on volatility, you do not need to cover all aspects, just focus on those occur in transcripts. \\

1. Earnings Surprises
    \begin{itemize}[itemsep=0pt]
        \item Magnitude of earnings per share (EPS) or revenue beats/misses relative to analyst estimates.
        \item Large surprises (positive or negative) often lead to dramatic market reactions due to adjustments in future expectations.
        \item A large miss might signal fundamental issues, triggering sell-offs, while a large beat might lead to euphoria, driving prices up significantly.
    \end{itemize}

2. Management Guidance
    \begin{itemize}[itemsep=0pt]
        \item Updates to forward-looking revenue, earnings, or margin expectations.
        \item  Upward guidance revision increases investor confidence, often reducing downside risk but amplifying upside potential, leading to volatility.
        \item Downward revisions heighten uncertainty and can trigger aggressive revaluation.
    \end{itemize}

3. Sentiment and Tone
    \begin{itemize}[itemsep=0pt]
        \item A confident tone with positive language like "record growth" or "strong demand" reduces perceived risk but might create higher expectations, increasing volatility if unmet.
        \item Defensive or overly cautious tone signals potential underlying issues, creating fear or speculation.
    \end{itemize}
... \\
... \\
... \\

Here is the earnings call transcripts: \textit{current\_earnings\_call\_transcripts}.

Earnings call transcripts ends. 

\end{tcolorbox}
\caption{Prompt with \textit{task-specific} strategy.}
\label{fig: Prompt for LLMs with task-specific strategy.}
\end{figure}

\paragraph{LLM Fine-grained Text}
We also leverage LLMs to extract signals and insights from vanilla earnings call transcripts. Specifically, we deploy two distinct prompt strategies:
    \begin{itemize}
        \item \textit{Summarization:} We prompt the LLMs to extract key points from the transcripts, focusing on different perspectives such as financial performance metrics, management commentary, and operational updates. The prompt template is illustrated in Figure~\ref{fig: Prompt for LLMs with summarization strategy.}.
        \item \textit{Task-Specific:} We provided the LLMs with the context of the post-earnings volatility prediction task, requiring them to generate insightful comments tailored to this objective. The prompt template is illustrated in Figure~\ref{fig: Prompt for LLMs with task-specific strategy.}.
    \end{itemize}
    
\subsection{Results for Transcripts-based Models.}
\label{Appendix Results for Transcripts-based Models.}
Due to the limited number of overlapping earnings in 2019 and 2020 within the DEC dataset\footnote{In real-world applications, it is typically possible to gather sufficient historical earnings data.}, which affects the suitability of the STPEV baseline, we present results only for the years 2021 to 2023 in Section~\ref{Evaluations on DEC Dataset}. Here, we provide the complete results on DEC, including three additional STPEV variants: STPEV(Median), STPEV(LR), and STPEV(MLP). The training-free \textit{median} implementation is similar to the \textit{mean} approach. The linear regression and MLP require training and parameter selection. To address this, we use cross-validation for each quarter, as earnings released in the same quarter share the same number of prior post-earnings volatility. 

As shown in Table~\ref{table: The Full Performance for DEC}, transcript-based models and STPEV(Mean) underperform STPEV(LR) during the early years of 2019 and 2020, where the limited number of previous earnings makes it challenging to capture the prior post-earnings volatility distribution. On the other hand, relatively complex implementations of STPEV, such as STPEV(LR) and STPEV(MLP), underperform simpler approaches like STPEV(Mean) and STPEV(Median) when sufficient previous same-ticker earnings are available.

\begin{table*}[ht]
\centering
\resizebox{\textwidth}{!}{%
\begin{tabular}{|c|c|c c c c c|c c c c c|c c c c c|c c c c c|c|}
\hline
\multirow{2}{*}{Year} &
\multirow{2}{*}{Model} & 
\multicolumn{5}{c|}{First Quarter} &
\multicolumn{5}{c|}{second Quarter} &
\multicolumn{5}{c|}{Third Quarter} &
\multicolumn{5}{c|}{Fourth Quarter} &
\multirow{2}{*}{Average}
 
\\
& & $\overline{MSE}$ & $MSE_3$ & $MSE_7$ & $MSE_{15}$ & $MSE_{30}$ 
& $\overline{MSE}$ & $MSE_3$ & $MSE_7$ & $MSE_{15}$ & $MSE_{30}$
& $\overline{MSE}$ & $MSE_3$ & $MSE_7$ & $MSE_{15}$ & $MSE_{30}$ 
& $\overline{MSE}$ & $MSE_3$ & $MSE_7$ & $MSE_{15}$ & $MSE_{30}$ & \\

\hline 
\multirow{12}{*}{2019} 
& Vanilla (Voyage) & - & - & - & - & - & 0.97 & 3.124 & 0.317 & 0.217 & 0.223 & 0.416 & 0.948 & 0.335 & 0.212 & 0.17 & 0.493 & 0.575 & 0.472 & 0.465 & 0.46 & 0.626  \\

& Vanilla (OpenAI) & - & - & - & - & - & 0.550 & 0.693 & 0.763 & 0.446 & 0.296 & \underline{0.258} & 0.563 & 0.245 & 0.130 & 0.092 & 0.266 & 0.466 & 0.263 & 0.196 & 0.140 & 0.358 \\

& Vanilla (Gecko) & - & - & - & - & - & 0.376 & 0.563 & 0.385 & 0.287 & 0.268 & 0.316 & 0.603 & 0.270 & 0.219 & 0.173 & 0.241 & 0.425 & 0.217 & 0.170 & 0.154 & 0.311 \\

& GPT4o (Summarization) & - & - & - & - & - & 0.317 & 0.596 & 0.398 & 0.168 & 0.104 & 0.279 & 0.585 & 0.272 & 0.155 & 0.104 & 0.247 & 0.433 & 0.227 & 0.171 & 0.159 & {0.281} \\

& GPT4o (Task-specific) & - & - & - & - & - & 0.328 & 0.606 & 0.416 & 0.180 & 0.109 & 0.285 & 0.605 & 0.278 & 0.154 & 0.104 & 0.231 & 0.407 & 0.214 & 0.162 & 0.142 & 0.281 \\

& Gemini (Summarization) & - & - & - & - & - & 0.349 & 0.646 & 0.444 & 0.193 & 0.112 & 0.277 & 0.585 & 0.268 & 0.153 & 0.102 & 0.230 & 0.409 & 0.218 & 0.157 & 0.138 & 0.285 \\

& Gemini (Task-specific) & - & - & - & - & - & 0.324 & 0.617 & 0.408 & 0.173 & 0.099 & 0.291 & 0.596 & 0.280 & 0.170 & 0.119 & 0.233 & 0.404 & 0.223 & 0.161 & 0.143 & 0.283 \\

& Random (All) & - & - & - & - & - & 0.326 & 0.603 & 0.376 & 0.197 & 0.129 & 0.358 & 0.623 & 0.328 & 0.288 & 0.191 & 0.276 & 0.448 & 0.256 & 0.206 & 0.192 & 0.320 \\

& Random (Ticker) & - & - & - & - & - & 0.878 & 1.733 & 0.389 & 0.442 & 0.947 & 0.322 & 0.599 & 0.310 & 0.225 & 0.154 & 0.228 & 0.396 & 0.214 & 0.158 & 0.145 & 0.476 \\

\cline{2-23}

& PEV(Mean) & - & - & - & - & - & \underline{0.266} & 0.522 & 0.288 & 0.149 & 0.107 & 0.294 & 0.574 & 0.279 & 0.188 & 0.133 & \underline{0.221} & 0.394 & 0.208 & 0.151 & 0.130 & \underline{0.260} \\

& STPEV(Mean) & - & - & - & - & - & 0.415 & 0.902 & 0.442 & 0.198 & 0.118 & 0.299 & 0.638 & 0.295 & 0.159 & 0.105 & 0.272 & 0.523 & 0.265 & 0.181 & 0.119 & 0.329 \\

& STPEV(Median) & - & - & - & - & - & 0.415 & 0.902 & 0.442 & 0.198 & 0.118 & 0.299 & 0.638 & 0.295 & 0.159 & 0.105 & 0.300 & 0.559 & 0.311 & 0.198 & 0.133 & 0.338 \\

& STPEV(LR) & - & - & - & - & - & \textbf{0.227} & 0.488 & 0.254 & 0.108 & 0.059 & \textbf{0.229} & 0.522 & 0.241 & 0.102 & 0.051 & \textbf{0.200} & 0.383 & 0.195 & 0.135 & 0.086 & \textbf{0.219} \\

& STPEV(MLP) & - & - & - & - & - & 0.648 & 1.122 & 0.665 & 0.459 & 0.348 & 0.305 & 0.671 & 0.330 & 0.150 & 0.068 & 0.259 & 0.504 & 0.259 & 0.174 & 0.099 & 0.404 \\

\hline\hline

\multirow{12}{*}{2020} & Vanilla (Voyage) & 0.782 & 0.695 & 0.422 & 0.662 & 1.351 & 0.594 & 0.791 & 0.438 & 0.4 & 0.748 & 0.574 & 0.879 & 0.47 & 0.501 & 0.445 & 0.335 & 0.616 & 0.259 & 0.24 & 0.225 & 0.571 \\

& Vanilla (OpenAI) & 0.783 & 0.744 & 0.447 & 0.646 & 1.295 & 0.339 & 0.637 & 0.321 & 0.235 & 0.162 & 0.328 & 0.684 & 0.245 & 0.209 & 0.177 & 0.325 & 0.540 & 0.275 & 0.284 & 0.200 & 0.444 \\

& Vanilla (Gecko) & 0.795 & 0.753 & 0.502 & 0.650 & 1.276 & 0.358 & 0.620 & 0.334 & 0.285 & 0.192 & 0.349 & 0.734 & 0.258 & 0.203 & 0.203 & 0.310 & 0.535 & 0.269 & 0.256 & 0.179 & 0.453 \\

& GPT4o (Summarization) & 0.781 & 0.701 & 0.470 & 0.652 & 1.302 & \underline{0.294} & 0.598 & 0.276 & 0.191 & 0.112 & 0.363 & 0.771 & 0.263 & 0.216 & 0.202 & \underline{0.268} & 0.482 & 0.222 & 0.220 & 0.147 & 0.427 \\

& GPT4o (Task-specific) & 0.787 & 0.733 & 0.479 & 0.648 & 1.286 & 0.342 & 0.619 & 0.329 & 0.261 & 0.159 & 0.311 & 0.690 & 0.209 & 0.169 & 0.176 & 0.280 & 0.508 & 0.225 & 0.234 & 0.153 & 0.430 \\

& Gemini (Summarization) & 0.787 & 0.679 & 0.463 & 0.666 & 1.340 & 0.306 & 0.610 & 0.294 & 0.207 & 0.114 & 0.369 & 0.769 & 0.271 & 0.239 & 0.196 & 0.277 & 0.483 & 0.223 & 0.233 & 0.170 & 0.435 \\

& Gemini (Task-specific) & 0.747 & 0.688 & 0.437 & 0.620 & 1.244 & 0.300 & 0.602 & 0.271 & 0.213 & 0.116 & 0.317 & 0.700 & 0.216 & 0.182 & 0.172 & 0.291 & 0.503 & 0.248 & 0.251 & 0.160 & 0.414 \\

& Random (All) & 0.909 & 0.842 & 0.563 & 0.751 & 1.481 & 0.393 & 0.625 & 0.359 & 0.345 & 0.244 & 0.372 & 0.760 & 0.299 & 0.237 & 0.191 & 0.351 & 0.546 & 0.288 & 0.340 & 0.232 & 0.506 \\

& Random (Ticker) & 0.810 & 0.756 & 0.496 & 0.680 & 1.308 & 0.437 & 0.694 & 0.413 & 0.379 & 0.262 & 0.351 & 0.760 & 0.271 & 0.204 & 0.171 & 0.346 & 0.551 & 0.304 & 0.310 & 0.221 & 0.486 \\

\cline{2-23}

& PEV(Mean) & 0.750 & 0.772 & 0.504 & 0.631 & 1.095 & 0.381 & 0.623 & 0.350 & 0.314 & 0.239 & 0.349 & 0.738 & 0.276 & 0.212 & 0.171 & 0.348 & 0.576 & 0.304 & 0.298 & 0.216 & 0.457 \\

& STPEV(Mean) & 0.817 & 0.725 & 0.477 & 0.695 & 1.370 & 0.438 & 0.786 & 0.383 & 0.346 & 0.237 & \textbf{0.269} & 0.685 & 0.196 & 0.112 & 0.083 & 0.311 & 0.536 & 0.275 & 0.281 & 0.151 & 0.459 \\

& STPEV(Median) & 0.814 & 0.735 & 0.480 & 0.684 & 1.358 & 0.495 & 0.798 & 0.441 & 0.396 & 0.345 & 0.283 & 0.700 & 0.201 & 0.128 & 0.103 & 0.337 & 0.519 & 0.295 & 0.318 & 0.215 & 0.482 \\

& STPEV(LR) & \textbf{0.422} & 0.664 & 0.370 & 0.453 & 0.200 & \textbf{0.277} & 0.614 & 0.222 & 0.148 & 0.126 & \underline{0.288} & 0.755 & 0.203 & 0.110 & 0.082 & \textbf{0.259} & 0.609 & 0.224 & 0.125 & 0.077 & \textbf{0.311} \\

& STPEV(MLP) & \underline{0.468} & 0.678 & 0.357 & 0.598 & 0.237 & 0.499 & 0.901 & 0.525 & 0.454 & 0.115 & 0.318 & 0.753 & 0.220 & 0.163 & 0.138 & 0.342 & 0.614 & 0.329 & 0.332 & 0.094 & \underline{0.407} \\

\hline\hline

\multirow{12}{*}{2021} & Vanilla (Voyage) & 0.432 & 0.566 & 0.363 & 0.325 & 0.473 &  0.378 & 0.644 & 0.366 & 0.274 & 0.227 & 0.371 & 0.669 & 0.426 & 0.263 & 0.127 & 0.238 & 0.512 & 0.192 & 0.128 & 0.12 & 0.355 \\

& Vanilla (OpenAI) & \underline{0.170} & 0.357 & 0.148 & 0.079 & 0.097 & 0.250 & 0.457 & 0.212 & 0.145 & 0.185 & 0.372 & 0.547 & 0.462 & 0.285 & 0.194 & \textbf{0.213} & 0.501 & 0.173 & 0.109 & 0.068 & 0.251 \\

& Vanilla (Gecko) & 0.200 & 0.419 & 0.191 & 0.104 & 0.087 & 0.269 & 0.464 & 0.245 & 0.176 & 0.189 & 0.350 & 0.523 & 0.377 & 0.291 & 0.211 & 0.253 & 0.535 & 0.223 & 0.161 & 0.094 & 0.268 \\

& GPT4o (Summarization) & 0.183 & 0.390 & 0.162 & 0.097 & 0.084 & 0.277 & 0.482 & 0.252 & 0.171 & 0.204 & 0.353 & 0.549 & 0.430 & 0.278 & 0.156 & 0.234 & 0.544 & 0.190 & 0.124 & 0.079 & 0.262 \\

& GPT4o (Task-specific) & 0.177 & 0.388 & 0.164 & 0.088 & 0.070 & \underline{0.246} & 0.444 & 0.214 & 0.145 & 0.180 & 0.357 & 0.515 & 0.428 & 0.295 & 0.189 & 0.242 & 0.537 & 0.197 & 0.145 & 0.088 & 0.255 \\

& Gemini (Summarization) & 0.175 & 0.356 & 0.153 & 0.083 & 0.106 & \underline{0.246} & 0.454 & 0.212 & 0.144 & 0.173 & 0.322 & 0.502 & 0.394 & 0.240 & 0.151 & 0.255 & 0.553 & 0.215 & 0.142 & 0.109 & 0.249 \\

& Gemini (Task-specific) & 0.176 & 0.384 & 0.159 & 0.094 & 0.067 & 0.249 & 0.435 & 0.210 & 0.156 & 0.197 & 0.347 & 0.503 & 0.407 & 0.286 & 0.190 & 0.248 & 0.548 & 0.215 & 0.140 & 0.088 & 0.255 \\

& Random (All) & 0.249 & 0.472 & 0.231 & 0.150 & 0.143 & 0.294 & 0.497 & 0.291 & 0.189 & 0.201 & 0.433 & 0.603 & 0.512 & 0.359 & 0.259 & 0.300 & 0.577 & 0.280 & 0.212 & 0.132 & 0.319 \\

& Random (Ticker) & 0.190 & 0.380 & 0.171 & 0.112 & 0.096 & 0.275 & 0.449 & 0.246 & 0.180 & 0.226 & 0.381 & 0.555 & 0.414 & 0.321 & 0.236 & 0.255 & 0.535 & 0.224 & 0.163 & 0.097 & 0.275 \\

\cline{2-23}

& PEV(Mean) & 0.216 & 0.433 & 0.209 & 0.115 & 0.105 & 0.271 & 0.451 & 0.239 & 0.184 & 0.209 & 0.405 & 0.580 & 0.429 & 0.342 & 0.270 & 0.288 & 0.568 & 0.260 & 0.199 & 0.127 & 0.295 \\

& STPEV(Mean) & \textbf{0.156} & 0.368 & 0.149 & 0.067 & 0.041 & 0.249 & 0.463 & 0.209 & 0.150 & 0.173 & 0.333 & 0.525 & 0.353 & 0.260 & 0.196 & \underline{0.222} & 0.536 & 0.177 & 0.114 & 0.062 & \underline{0.240} \\

& STPEV(Median) & 0.174 & 0.406 & 0.168 & 0.077 & 0.046 & 0.241 & 0.485 & 0.214 & 0.128 & 0.138 & 0.340 & 0.594 & 0.361 & 0.241 & 0.164 & 0.243 & 0.580 & 0.189 & 0.129 & 0.073 & 0.250 \\

& STPEV(LR) & 0.185 & 0.443 & 0.180 & 0.073 & 0.045 & \textbf{0.191} & 0.412 & 0.170 & 0.114 & 0.069 & \textbf{0.277} & 0.602 & 0.249 & 0.158 & 0.099 & 0.248 & 0.548 & 0.235 & 0.139 & 0.072 & \textbf{0.225} \\

& STPEV(MLP) & 0.230 & 0.537 & 0.194 & 0.125 & 0.063 & 0.402 & 0.458 & 0.182 & 0.142 & 0.827 & \underline{0.315} & 0.537 & 0.362 & 0.161 & 0.200 & 0.268 & 0.527 & 0.220 & 0.167 & 0.159 & 0.304 \\

\hline\hline
 
\multirow{12}{*}{2022} & Vanilla (Voyage) & 0.247 & 0.522 & 0.231 & 0.135 & 0.101 &  0.237 & 0.535 & 0.174 & 0.132 & 0.108 &  0.33 & 0.713 & 0.29 & 0.184 & 0.131 &  0.241 & 0.543 & 0.177 & 0.135 & 0.11 & 0.264 \\

& Vanilla (OpenAI) & \textbf{0.258} & 0.523 & 0.253 & 0.160 & 0.095 & 0.354 & 0.671 & 0.251 & 0.260 & 0.235 & 0.261 & 0.622 & 0.237 & 0.112 & 0.072 & \underline{0.256} & 0.585 & 0.195 & 0.156 & 0.088 & 0.282 \\

& Vanilla (Gecko) & 0.302 & 0.603 & 0.293 & 0.188 & 0.125 & 0.353 & 0.659 & 0.325 & 0.238 & 0.188 & 0.265 & 0.608 & 0.234 & 0.134 & 0.083 & 0.274 & 0.588 & 0.228 & 0.178 & 0.101 & 0.298 \\

& GPT4o (Summarization) & 0.332 & 0.641 & 0.311 & 0.225 & 0.149 & 0.317 & 0.656 & 0.273 & 0.174 & 0.164 & \textbf{0.239} & 0.586 & 0.195 & 0.107 & 0.069 & 0.291 & 0.629 & 0.217 & 0.201 & 0.118 & 0.295 \\

& GPT4o (Task-specific) & 0.309 & 0.585 & 0.286 & 0.220 & 0.147 & \underline{0.307} & 0.618 & 0.196 & 0.193 & 0.220 & \underline{0.243} & 0.578 & 0.219 & 0.110 & 0.065 & 0.275 & 0.599 & 0.209 & 0.194 & 0.099 & 0.284 \\

& Gemini (Summarization) & 0.297 & 0.591 & 0.269 & 0.208 & 0.119 & 0.348 & 0.660 & 0.240 & 0.244 & 0.246 & 0.252 & 0.605 & 0.220 & 0.109 & 0.073 & \textbf{0.250} & 0.573 & 0.180 & 0.151 & 0.094 & 0.287 \\

& Gemini (Task-specific) & 0.291 & 0.599 & 0.276 & 0.169 & 0.120 & 0.314 & 0.656 & 0.254 & 0.187 & 0.157 & \underline{0.243} & 0.584 & 0.216 & 0.109 & 0.065 & 0.267 & 0.604 & 0.191 & 0.174 & 0.098 & 0.279 \\

& Random (All) & 0.316 & 0.614 & 0.326 & 0.203 & 0.121 & 0.410 & 0.746 & 0.390 & 0.291 & 0.213 & 0.324 & 0.674 & 0.298 & 0.195 & 0.131 & 0.324 & 0.597 & 0.304 & 0.231 & 0.163 & 0.343 \\

& Random (Ticker) & \underline{0.270} & 0.553 & 0.269 & 0.159 & 0.098 & 0.308 & 0.609 & 0.235 & 0.223 & 0.163 & 0.255 & 0.610 & 0.230 & 0.113 & 0.068 & 0.285 & 0.602 & 0.242 & 0.187 & 0.108 & 0.279 \\

\cline{2-23}

& PEV(Mean) & 0.326 & 0.619 & 0.326 & 0.215 & 0.146 & 0.380 & 0.719 & 0.343 & 0.272 & 0.185 & 0.310 & 0.647 & 0.285 & 0.185 & 0.121 & 0.316 & 0.618 & 0.283 & 0.220 & 0.143 & 0.333 \\

& STPEV(Mean) & \underline{0.270} & 0.584 & 0.245 & 0.152 & 0.099 & 0.310 & 0.640 & 0.270 & 0.201 & 0.129 & \underline{0.243} & 0.592 & 0.219 & 0.104 & 0.057 & 0.278 & 0.599 & 0.236 & 0.183 & 0.095 & \textbf{0.275} \\

& STPEV(Median) & 0.268 & 0.544 & 0.245 & 0.164 & 0.119 & 0.320 & 0.623 & 0.273 & 0.228 & 0.154 & 0.237 & 0.576 & 0.210 & 0.101 & 0.061 & 0.282 & 0.592 & 0.233 & 0.199 & 0.102 & \underline{0.277} \\

& STPEV(LR) & 0.305 & 0.739 & 0.273 & 0.145 & 0.063 & \textbf{0.251} & 0.759 & 0.140 & 0.075 & 0.030 & 0.292 & 0.723 & 0.258 & 0.120 & 0.068 & 0.264 & 0.690 & 0.195 & 0.107 & 0.065 & {0.278} \\

& STPEV(MLP) & 0.479 & 1.133 & 0.264 & 0.194 & 0.325 & 0.354 & 0.777 & 0.210 & 0.272 & 0.157 & 0.486 & 0.910 & 0.271 & 0.699 & 0.066 & 0.716 & 1.090 & 0.821 & 0.800 & 0.155 & 0.509 \\

\hline\hline

\multirow{12}{*}{2023} & Vanilla (Voyage) & 0.55 & 0.945 & 0.476 & 0.385 & 0.392 &   0.283 & 0.648 & 0.243 & 0.153 & 0.087 &   0.23 & 0.505 & 0.223 & 0.117 & 0.077 & 0.24 & 0.526 & 0.213 & 0.124 & 0.096 &  0.326\\

& Vanilla (OpenAI) & 0.268 & 0.663 & 0.197 & 0.109 & 0.104 & 0.274 & 0.643 & 0.235 & 0.134 & 0.084 & 0.226 & 0.439 & 0.249 & 0.129 & 0.088 & 0.258 & 0.557 & 0.222 & 0.142 & 0.111 & 0.257 \\

& Vanilla (Gecko) & 0.257 & 0.659 & 0.189 & 0.104 & 0.076 & 0.266 & 0.619 & 0.226 & 0.132 & 0.089 & 0.215 & 0.408 & 0.213 & 0.126 & 0.113 & \textbf{0.229} & 0.473 & 0.208 & 0.131 & 0.105 & \underline{0.242} \\

& GPT4o (Summarization) & 0.262 & 0.642 & 0.202 & 0.108 & 0.097 & 0.266 & 0.634 & 0.233 & 0.121 & 0.076 & 0.220 & 0.405 & 0.226 & 0.144 & 0.103 & 0.250 & 0.542 & 0.214 & 0.145 & 0.098 & 0.249 \\

& GPT4o (Task-specific) & 0.249 & 0.634 & 0.188 & 0.097 & 0.078 & 0.262 & 0.621 & 0.221 & 0.125 & 0.080 & 0.220 & 0.418 & 0.209 & 0.142 & 0.110 & 0.253 & 0.542 & 0.220 & 0.140 & 0.110 & 0.246 \\

& Gemini (Summarization) & 0.262 & 0.629 & 0.194 & 0.113 & 0.114 & 0.260 & 0.624 & 0.213 & 0.129 & 0.074 & \textbf{0.210} & 0.415 & 0.224 & 0.119 & 0.083 & \underline{0.238} & 0.523 & 0.210 & 0.127 & 0.092 & 0.243 \\

& Gemini (Task-specific) & 0.262 & 0.637 & 0.204 & 0.115 & 0.092 & 0.269 & 0.622 & 0.228 & 0.136 & 0.088 & \textbf{0.210} & 0.408 & 0.205 & 0.126 & 0.102 & 0.244 & 0.528 & 0.211 & 0.133 & 0.105 & 0.246 \\

& Random (All) & 0.317 & 0.729 & 0.248 & 0.163 & 0.130 & 0.352 & 0.752 & 0.290 & 0.213 & 0.152 & 0.280 & 0.481 & 0.265 & 0.193 & 0.182 & 0.279 & 0.566 & 0.251 & 0.155 & 0.143 & 0.307 \\

& Random (Ticker) & \underline{0.247} & 0.633 & 0.182 & 0.095 & 0.080 & \underline{0.255} & 0.596 & 0.208 & 0.130 & 0.088 & 0.228 & 0.438 & 0.212 & 0.133 & 0.130 & \underline{0.238} & 0.513 & 0.203 & 0.124 & 0.110 & \underline{0.242} \\

\cline{2-23}

& PEV(Mean) & 0.309 & 0.725 & 0.236 & 0.150 & 0.124 & 0.330 & 0.723 & 0.279 & 0.186 & 0.134 & 0.262 & 0.463 & 0.249 & 0.172 & 0.166 & 0.278 & 0.584 & 0.245 & 0.148 & 0.134 & 0.295 \\

& STPEV(Mean) & \textbf{0.239} & 0.611 & 0.180 & 0.093 & 0.074 & \textbf{0.253} & 0.601 & 0.209 & 0.122 & 0.081 & 0.227 & 0.432 & 0.215 & 0.132 & 0.130 & 0.246 & 0.520 & 0.215 & 0.133 & 0.118 & \underline{0.242} \\

& STPEV(Median) & 0.240 & 0.609 & 0.178 & 0.095 & 0.078 & 0.262 & 0.650 & 0.214 & 0.111 & 0.075 & 0.214 & 0.405 & 0.210 & 0.124 & 0.119 & 0.240 & 0.494 & 0.222 & 0.136 & 0.110 & \textbf{0.239} \\

& STPEV(LR) & 0.332 & 0.851 & 0.272 & 0.127 & 0.078 & 0.290 & 0.714 & 0.231 & 0.142 & 0.071 & 0.293 & 0.542 & 0.342 & 0.175 & 0.112 & 0.333 & 0.767 & 0.311 & 0.157 & 0.097 & 0.312 \\

& STPEV(MLP) & 0.820 & 1.263 & 1.280 & 0.394 & 0.345 & 0.565 & 1.031 & 0.605 & 0.431 & 0.195 & 0.397 & 0.446 & 0.275 & 0.557 & 0.311 & 0.490 & 0.625 & 0.526 & 0.469 & 0.341 & 0.568 \\

\hline

\end{tabular}
}
\caption{The overall performance on DEC across transcript-based models and different PEV and STPEV variants.}
\label{table: The Full Performance for DEC}
\end{table*}

\subsection{Pre-earnings Volatility Series Prediction.}
Since volatility prediction is a type of time-series forecasting problem, we follow KeFVP~\cite{niu-etal-2023-kefvp} and predict volatility using \textit{pre-earnings volatility series} (with a window size of 22\footnote{In the United States, the number of trading days in a month typically ranges from 20 to 22 days.}) through various time-series forecasting models\footnote{\url{https://github.com/thuml/Time-Series-Library}}, including DLinear \cite{zeng2023transformers}, TSMixer \cite{chen2023tsmixerallmlparchitecturetime}, TimeNet \cite{wu2022timesnet}, and FEDformer \cite{zhou2022fedformer}. Table~\ref{table: The Performance on TSF DEC} compares these time-series forecasting models with STPEV(Mean). We observe that simpler models, such as DLinear and TSMixer, outperform other forecasting models, though their performance and that of STPEV(Mean) vary across different years. This suggests that using pre-earnings volatility time series is also an effective approach for post-earnings volatility prediction. Future research could explore combining \textit{pre-earnings} and textit{post-earnings} volatility series to achieve improved prediction accuracy.

\begin{table*}[ht]
\centering
\resizebox{\textwidth}{!}{%
\begin{tabular}{|cc|c c c c c|c c c c c|c c c c c|c c c c c|c|}
\hline
\multirow{2}{*}{Year} &
\multirow{2}{*}{Model} & 
\multicolumn{5}{c|}{First Quarter} &
\multicolumn{5}{c|}{second Quarter} &
\multicolumn{5}{c|}{Third Quarter} &
\multicolumn{5}{c|}{Fourth Quarter} &
\multirow{2}{*}{ Average}
\\
& & $\overline{MSE}$ & $MSE_3$ & $MSE_7$ & $MSE_{15}$ & $MSE_{30}$ 
& $\overline{MSE}$ & $MSE_3$ & $MSE_7$ & $MSE_{15}$ & $MSE_{30}$
& $\overline{MSE}$ & $MSE_3$ & $MSE_7$ & $MSE_{15}$ & $MSE_{30}$ 
& $\overline{MSE}$ & $MSE_3$ & $MSE_7$ & $MSE_{15}$ & $MSE_{30}$  & \\

\hline 
\multirow{5}{*}{2019} & DLinear & - & - & - & - & - & 0.438 & 0.500 & 0.370 & 0.435 & 0.449 & \underline{0.322} & 0.567 & 0.284 & 0.258 & 0.178 & 0.235 & 0.456 & 0.220 & 0.147 & 0.117 & 0.332 \\

& TSMixer & - & - & - & - & - & \underline{0.413} & 0.535 & 0.244 & 0.278 & 0.595 & 0.330 & 0.554 & 0.299 & 0.392 & 0.076 & \textbf{0.224} & 0.399 & 0.220 & 0.148 & 0.130 & \textbf{0.322} \\

& TimesNet & - & - & - & - & - & \textbf{0.394} & 0.628 & 0.331 & 0.410 & 0.208 & 0.358 & 0.593 & 0.366 & 0.301 & 0.171 & \underline{0.230} & 0.395 & 0.268 & 0.153 & 0.102 & \underline{0.327} \\

& FEDformer & - & - & - & - & - & 0.706 & 0.743 & 0.733 & 0.982 & 0.366 & 0.325 & 0.724 & 0.270 & 0.177 & 0.128 & 0.267 & 0.459 & 0.283 & 0.178 & 0.149 & 0.433 \\

& STPEV(Mean) & - & - & - & - & - & 0.415 & 0.902 & 0.442 & 0.198 & 0.118 & \textbf{0.299} & 0.638 & 0.295 & 0.159 & 0.105 & 0.272 & 0.523 & 0.265 & 0.181 & 0.119 & 0.329 \\

\hline 
\multirow{5}{*}{2020} & DLinear & 0.929 & 0.602 & 0.426 & 0.709 & 1.980 & 1.158 & 1.216 & 0.942 & 1.064 & 1.411 & 0.433 & 0.803 & 0.326 & 0.289 & 0.312 & \textbf{0.246} & 0.489 & 0.199 & 0.177 & 0.119 & 0.691 \\

& TSMixer & 0.920 & 0.677 & 0.439 & 0.606 & 1.956 & \underline{0.595} & 0.784 & 0.497 & 0.505 & 0.594 & \underline{0.381} & 0.784 & 0.301 & 0.229 & 0.211 & \underline{0.273} & 0.504 & 0.208 & 0.232 & 0.147 & \underline{0.542} \\

& TimesNet & 0.938 & 0.682 & 0.603 & 0.628 & 1.839 & 1.081 & 1.233 & 0.829 & 1.031 & 1.233 & 0.422 & 0.954 & 0.302 & 0.231 & 0.201 & 0.331 & 0.518 & 0.275 & 0.280 & 0.250 & 0.693 \\

& FEDformer & \textbf{0.772} & 0.654 & 0.322 & 0.499 & 1.613 & 0.912 & 1.251 & 0.903 & 0.714 & 0.781 & 0.442 & 0.883 & 0.329 & 0.293 & 0.264 & 0.318 & 0.610 & 0.230 & 0.211 & 0.220 & 0.611 \\

& STPEV(Mean) & \underline{0.817} & 0.725 & 0.477 & 0.695 & 1.370 & \textbf{0.438} & 0.786 & 0.383 & 0.346 & 0.237 & \textbf{0.269} & 0.685 & 0.196 & 0.112 & 0.083 & 0.311 & 0.536 & 0.275 & 0.281 & 0.151 & \textbf{0.459} \\

\hline 
\multirow{5}{*}{2021} & DLinear & 0.192 & 0.434 & 0.188 & 0.086 & 0.061 & \textbf{0.215} & 0.422 & 0.190 & 0.124 & 0.123 & \textbf{0.256} & 0.498 & 0.267 & 0.161 & 0.097 & 0.248 & 0.549 & 0.207 & 0.146 & 0.092 & \textbf{0.228} \\

& TSMixer & \underline{0.191} & 0.434 & 0.193 & 0.083 & 0.053 & 0.226 & 0.401 & 0.162 & 0.141 & 0.199 & \underline{0.275} & 0.461 & 0.293 & 0.208 & 0.140 & \underline{0.246} & 0.542 & 0.207 & 0.149 & 0.087 & \underline{0.235} \\

& TimesNet & 0.221 & 0.409 & 0.226 & 0.170 & 0.078 & \underline{0.223} & 0.409 & 0.229 & 0.147 & 0.106 & 0.290 & 0.490 & 0.285 & 0.250 & 0.136 & 0.259 & 0.520 & 0.225 & 0.169 & 0.121 & 0.248 \\

& FEDformer & 0.254 & 0.463 & 0.235 & 0.175 & 0.141 & 0.267 & 0.472 & 0.233 & 0.187 & 0.177 & 0.305 & 0.553 & 0.292 & 0.203 & 0.173 & 0.272 & 0.540 & 0.277 & 0.175 & 0.095 & 0.274 \\

& STPEV(Mean) & \textbf{0.156} & 0.368 & 0.149 & 0.067 & 0.041 & 0.249 & 0.463 & 0.209 & 0.150 & 0.173 & 0.333 & 0.525 & 0.353 & 0.260 & 0.196 & \textbf{0.222} & 0.536 & 0.177 & 0.114 & 0.062 & 0.240 \\

\hline 
\multirow{5}{*}{2022} & DLinear & \textbf{0.244} & 0.553 & 0.212 & 0.133 & 0.079 & \textbf{0.208} & 0.490 & 0.179 & 0.112 & 0.052 & 0.304 & 0.597 & 0.253 & 0.198 & 0.169 & 0.235 & 0.555 & 0.185 & 0.111 & 0.087 & \underline{0.248} \\

& TSMixer & \underline{0.246} & 0.560 & 0.214 & 0.121 & 0.090 & 0.229 & 0.484 & 0.196 & 0.157 & 0.080 & \underline{0.257} & 0.570 & 0.216 & 0.140 & 0.100 & \textbf{0.229} & 0.527 & 0.198 & 0.113 & 0.079 & \textbf{0.240} \\

& TimesNet & 0.266 & 0.562 & 0.233 & 0.155 & 0.114 & 0.239 & 0.499 & 0.185 & 0.167 & 0.106 & 0.270 & 0.601 & 0.221 & 0.136 & 0.123 & \underline{0.231} & 0.516 & 0.198 & 0.099 & 0.112 & 0.252 \\

& FEDformer & 0.282 & 0.627 & 0.281 & 0.142 & 0.078 & \underline{0.218} & 0.519 & 0.147 & 0.110 & 0.097 & 0.296 & 0.604 & 0.251 & 0.154 & 0.176 & 0.246 & 0.559 & 0.200 & 0.128 & 0.098 & 0.261 \\

& STPEV(Mean) & 0.270 & 0.584 & 0.245 & 0.152 & 0.099 & 0.310 & 0.640 & 0.270 & 0.201 & 0.129 & \textbf{0.243} & 0.592 & 0.219 & 0.104 & 0.057 & 0.278 & 0.599 & 0.236 & 0.183 & 0.095 & 0.275 \\

\hline 
\multirow{5}{*}{2023} & DLinear & \underline{0.254} & 0.667 & 0.171 & 0.106 & 0.072 & 0.336 & 0.789 & 0.291 & 0.163 & 0.101 & 0.254 & 0.519 & 0.259 & 0.147 & 0.089 & 0.299 & 0.666 & 0.279 & 0.157 & 0.094 & 0.286 \\

& TSMixer & 0.255 & 0.671 & 0.183 & 0.102 & 0.064 & \underline{0.304} & 0.724 & 0.262 & 0.142 & 0.088 & \textbf{0.223} & 0.456 & 0.223 & 0.122 & 0.090 & \underline{0.265} & 0.592 & 0.241 & 0.137 & 0.091 & \underline{0.262} \\

& TimesNet & 0.268 & 0.652 & 0.189 & 0.117 & 0.116 & 0.365 & 0.838 & 0.325 & 0.169 & 0.128 & 0.296 & 0.599 & 0.302 & 0.177 & 0.107 & 0.341 & 0.741 & 0.322 & 0.182 & 0.119 & 0.318 \\

& FEDformer & 0.273 & 0.702 & 0.186 & 0.117 & 0.087 & 0.363 & 0.863 & 0.309 & 0.168 & 0.113 & 0.282 & 0.571 & 0.294 & 0.163 & 0.101 & 0.316 & 0.712 & 0.281 & 0.158 & 0.115 & 0.309 \\

& STPEV(Mean) & \textbf{0.239} & 0.611 & 0.180 & 0.093 & 0.074 & \textbf{0.253} & 0.601 & 0.209 & 0.122 & 0.081 & \underline{0.227} & 0.432 & 0.215 & 0.132 & 0.130 & \textbf{0.246} & 0.520 & 0.215 & 0.133 & 0.118 & \textbf{0.242} \\

\hline
\end{tabular}
}
\caption{The overall performance on DEC using pre-earnings volatility series. }
\label{table: The Performance on TSF DEC}
\end{table*}

\begin{figure*}[!htb]  
    \centering
    \begin{subfigure}{.48\textwidth}
        \centering
        \includegraphics[width=\linewidth]{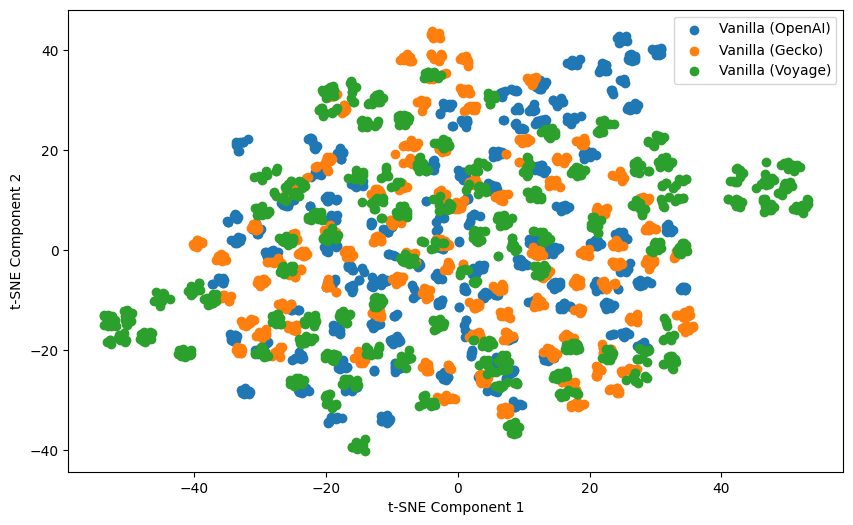}
        \caption{Three embeddings for vanilla transcripts.}
        \label{Three embeddings for vanilla transcripts.}
       
    \end{subfigure}%
    \hfill
    \begin{subfigure}{.48\textwidth}
        \centering
        \includegraphics[width=\linewidth]{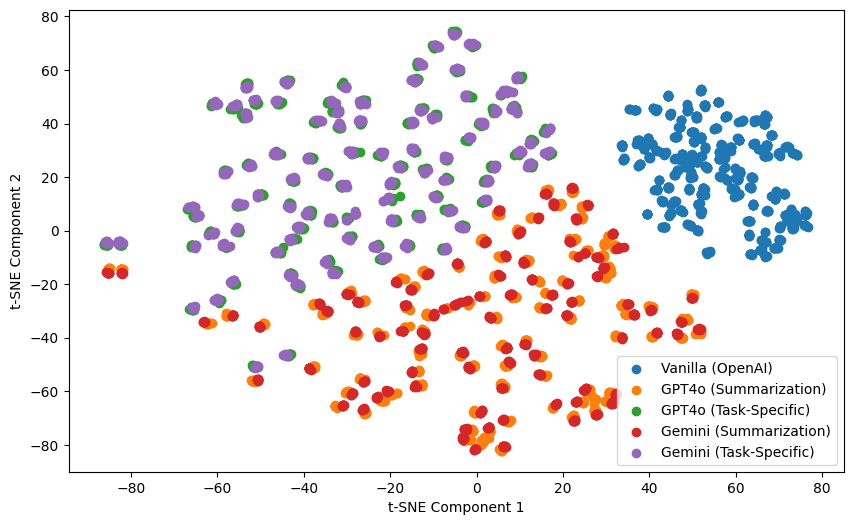}
        \caption{Vanilla transcripts and four LLMs fine-grained texts.}
        \label{Vanilla transcripts and four LLMs fine-grained texts.}
     
    \end{subfigure}%

    \caption{Clustering results for different embeddding models and texts.}
    \label{Figure: Clustering results for different embeddding models and texts.}
\end{figure*}

\subsection{Clustering of Different Transcripts Embeddings}
\label{Appendix Clustering among Different Text}
We compare three embedding models\footnote{Specifically, OpenAI text-embedding-3-large model, Text-embedding-005 model, and Voyage-Finance-2 model.}. for vanilla transcripts by sampling 10 earnings from each ticker, resulting in 900 earnings. We then perform K-means clustering on 2,700 embeddings after t-SNE dimensionality reduction. Figure~\ref{Three embeddings for vanilla transcripts.} displays the clustering results, showing that embeddings generated by the three models tend to mix with one another. Additionally, all models exhibit ticker groupings, reflecting the inherent similarity among transcripts from the same company.

We further compare the embeddings\footnote{The OpenAI text-embedding-3-large model is used.} of vanilla transcripts with those of four types of LLM fine-grained insights (two strategies and two LLMs). Figure~\ref{Vanilla transcripts and four LLMs fine-grained texts.} presents the clustering results, showing that \textit{vanilla transcripts}, \textit{Summarization}, and textit{Task-Specific} texts are distinctly separated from others. In contrast, the two \textit{LLMs—GPT4o-2024-08-06} and \textit{Gemini-1.5-Flash}—struggle to differentiate from each other. Furthermore, secondary clusters emerge within the three main clusters, reflecting ticker-specific groupings.

\subsection{Group Similarity Comparison}
\label{Appendix Similarity Comparison}

\begin{figure*}[!htb]  
    \centering
    \begin{subfigure}{.24\textwidth}
        \centering
        \includegraphics[width=\linewidth]{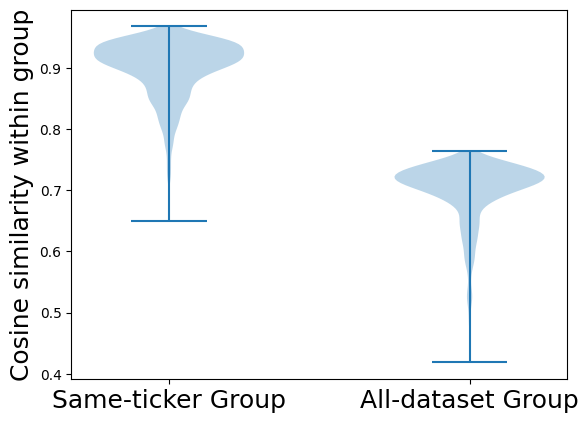}
        \caption{Vanilla (OpenAI)}
       
    \end{subfigure}%
    \hfill
    \begin{subfigure}{.24\textwidth}
        \centering
        \includegraphics[width=\linewidth]{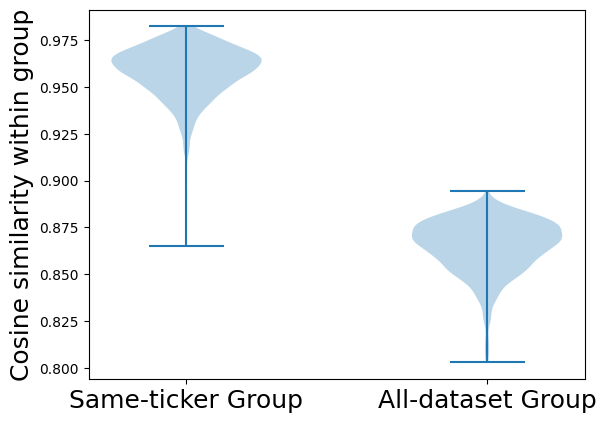}
        \caption{Vanilla (Gecko)}
     
    \end{subfigure}%
    \hfill
    \begin{subfigure}{.24\textwidth}
        \centering
        \includegraphics[width=\linewidth]{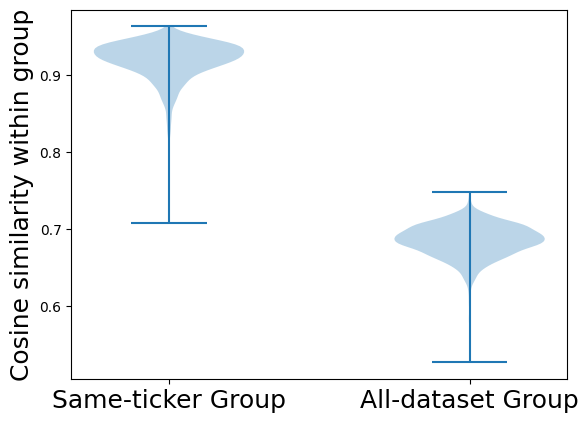}
        \caption{GPT4o (Summarization)}
        \label{fig:ec}
    \end{subfigure}%
    \hfill
    \begin{subfigure}{.24\textwidth}
        \centering
        \includegraphics[width=\linewidth]{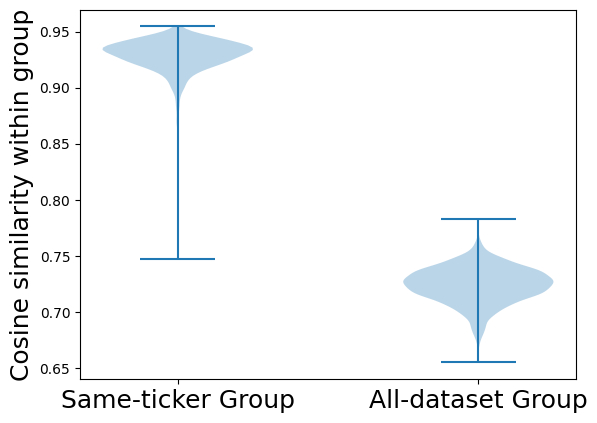}
        \caption{GPT4o (Task-Specific)}
        \label{fig:gemini_vol2}
    \end{subfigure}%
    \hfill

    \par\bigskip 

    
    \begin{subfigure}{.24\textwidth}
        \centering
        \includegraphics[width=\linewidth]{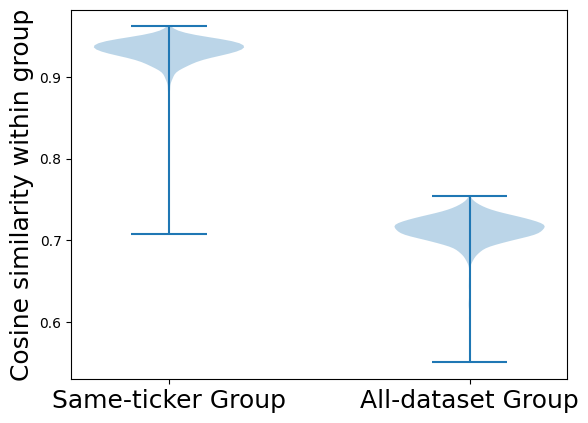}
        \caption{Gemini (Summarization)}
        \label{fig:gemini_sum}
    \end{subfigure}%
    \hfill
    \begin{subfigure}{.24\textwidth}
        \centering
        \includegraphics[width=\linewidth]{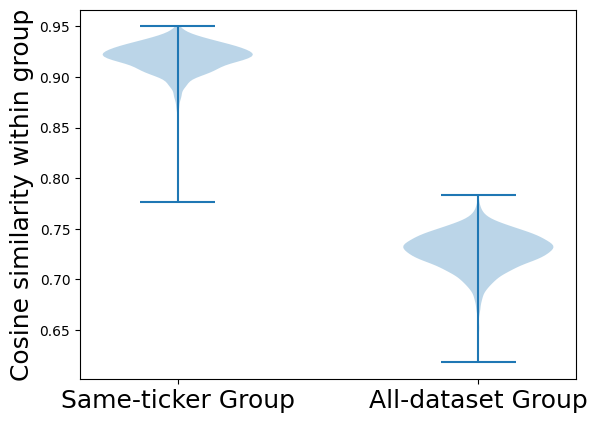}
        \caption{Gemini (Task Specific)}
        \label{fig:gemini_vol1}
    \end{subfigure}%
    \hfill
    \begin{subfigure}{.24\textwidth}
        \centering
        \includegraphics[width=\linewidth]{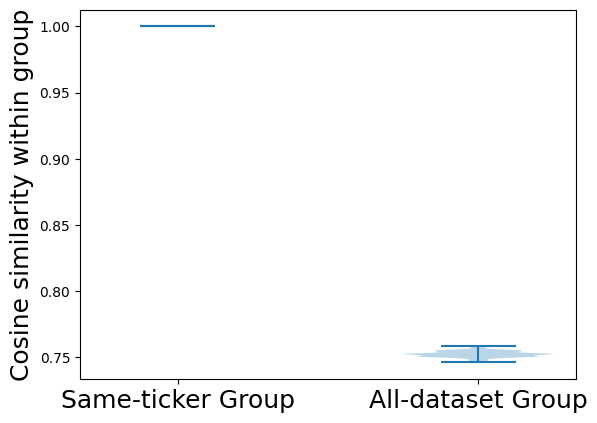}
        \caption{Random Ticker}
        \label{fig:random_all}
    \end{subfigure}%
    \hfill
    \begin{subfigure}{.24\textwidth}
        \centering
        \includegraphics[width=\linewidth]{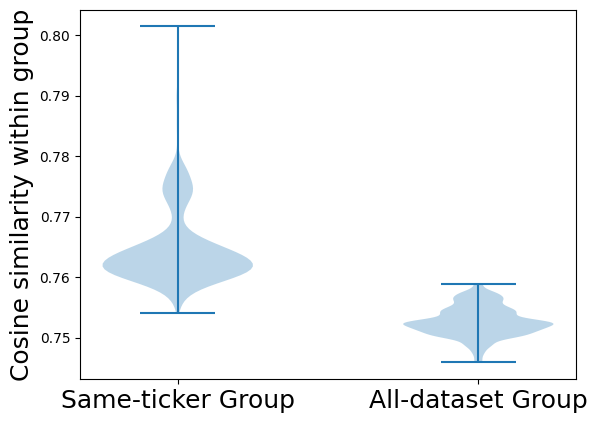}
        \caption{Random All}
        \label{fig:random_subset}
    \end{subfigure}
    
    \caption{Cosine similarity comparison between same-ticker group and all-dataset group.}
    \label{Figure: Cosine Similarity Same-ticker group and All-dataset group}
\end{figure*}

To evaluate how transcript representations correlate with other examples from the same-ticker group (comprising 1 ticker and 19 earnings) and the all-dataset group (comprising all 90 tickers and 1799 earnings), we compute the cosine similarity of each earnings record with both groups. This analysis is performed across different transcript representations, including two vanilla embeddings and four fine-grained LLM embeddings. As shown in Table~\ref{Figure: Cosine Similarity Same-ticker group and All-dataset group}, the cosine similarity for the same-ticker group is significantly higher than that for the all-dataset group, except for Random(All) representations, where we intentionally remove the ticker identity. This finding indicates that transcripts from the same ticker, which implicitly contain the ticker identity, are more similar than those from different tickers.

\begin{figure*}[!htb]  
    \centering
        \centering        \includegraphics[scale=0.3]{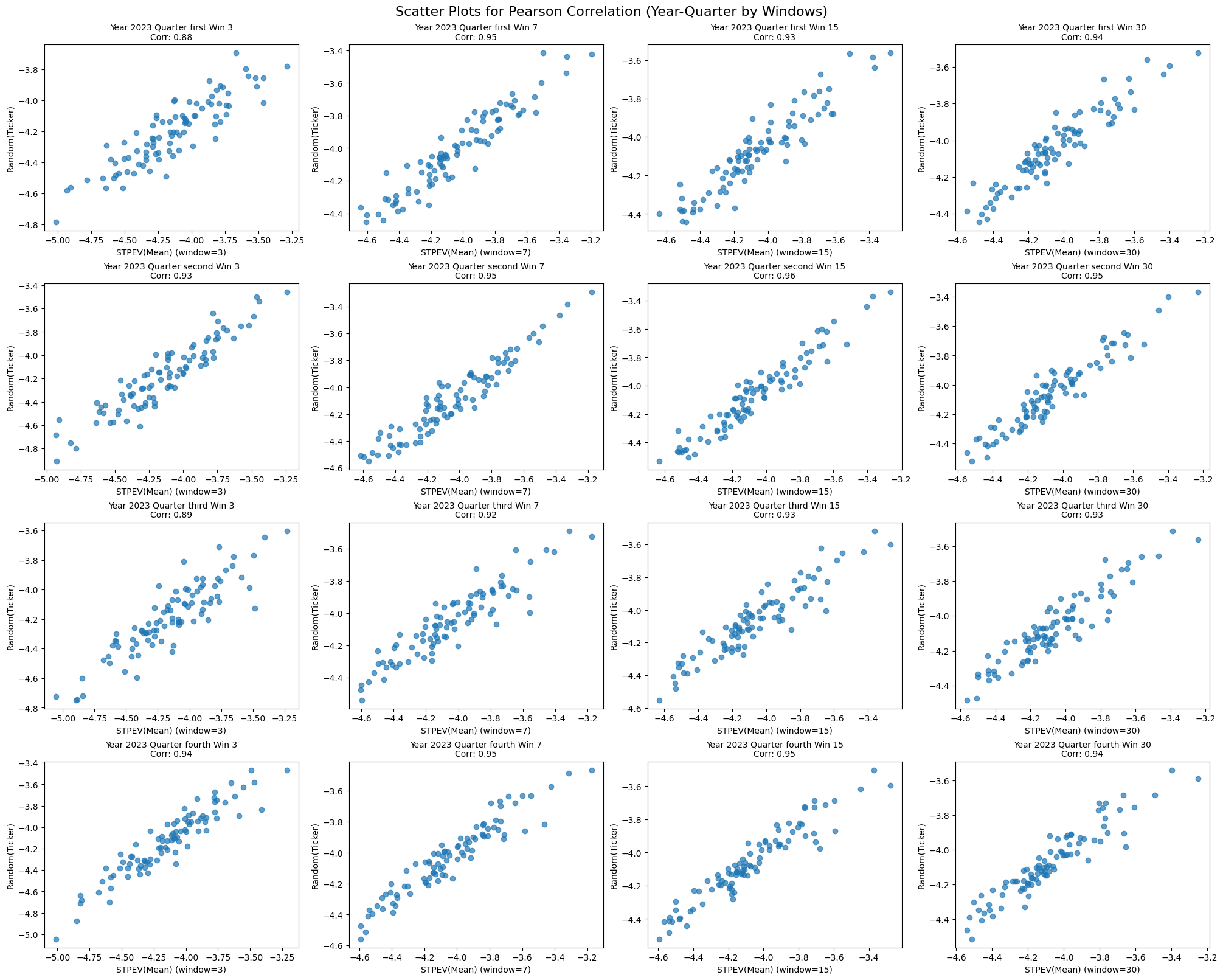}
    \caption{Predictions by Random(Ticker) - Predictions by STPEV(Mean) model for year 2023.}
    \label{fig:Pearson Random(Ticker)}
\end{figure*}


\subsection{Predictions Correlation Analysis}
\label{Appendix Predictions Correlation}
We also calculate and compare the Pearson correlation coefficients between the predictions of various transcript-based models and those of STPEV(Mean) across different windows and quarters. From Table~\ref{table: Pearson Vanilla} to Table~\ref{table: Pearson Random(Ticker)}, We observe strong linear relationships between the predictions generated by transcript-based models and STPEV(Mean), typically beginning from the year 2021. Figure~\ref{fig:Pearson Random(Ticker)} further visualizes this relationship for the year 2023, comparing predictions from the Random(Ticker) model with those from STPEV(Mean). The points cluster closely around the line  y = x , highlighting a strong correlation.

\begin{table*}[ht]
\centering
\resizebox{\textwidth}{!}{%
\begin{tabular}{|c|c c c c c|c c c c c|c c c c c|c c c c c|c|}
\hline
\multirow{2}{*}{Year} &
\multicolumn{5}{c|}{First Quarter} &
\multicolumn{5}{c|}{second Quarter} &
\multicolumn{5}{c|}{Third Quarter} &
\multicolumn{5}{c|}{Fourth Quarter} &
\multirow{2}{*}{Yearly Average}
\\
& $\overline{Coef}$ & $Coef_3$ & $Coef_7$ & $Coef_{15}$ & $Coef_{30}$ 
& $\overline{Coef}$ & $Coef_3$ & $Coef_7$ & $Coef_{15}$ & $Coef_{30}$
& $\overline{Coef}$ & $Coef_3$ & $Coef_7$ & $Coef_{15}$ & $Coef_{30}$ 
& $\overline{Coef}$ & $Coef_3$ & $Coef_7$ & $Coef_{15}$ & $Coef_{30}$  & \\

\hline
2019 & - & - & - & - & - & 0.14 & 0.169 & 0.146 & 0.125 & 0.121 & 0.477 & 0.483 & 0.456 & 0.472 & 0.496 & 0.415 & 0.469 & 0.431 & 0.39 & 0.37  &  0.556
\\
\hline
2020 & 0.556 & 0.601 & 0.531 & 0.548 & 0.546 & 0.521 & 0.58 & 0.583 & 0.482 & 0.437 & 0.555 & 0.603 & 0.549 & 0.533 & 0.537 & 0.62 & 0.651 & 0.632 & 0.633 & 0.563  &  0.767
\\
\hline
2021 & 0.767 & 0.755 & 0.801 & 0.766 & 0.745 & 0.8 & 0.788 & 0.807 & 0.812 & 0.794 & 0.801 & 0.803 & 0.789 & 0.831 & 0.781 & 0.845 & 0.782 & 0.87 & 0.862 & 0.866  &  0.87
\\
\hline
2022 & 0.87 & 0.857 & 0.866 & 0.873 & 0.884 & 0.81 & 0.794 & 0.794 & 0.842 & 0.809 & 0.853 & 0.82 & 0.863 & 0.873 & 0.856 & 0.866 & 0.835 & 0.885 & 0.877 & 0.866  &  0.873
\\
\hline
2023 & 0.873 & 0.843 & 0.897 & 0.89 & 0.864 & 0.869 & 0.847 & 0.877 & 0.874 & 0.878 & 0.838 & 0.778 & 0.835 & 0.864 & 0.877 & 0.866 & 0.852 & 0.859 & 0.87 & 0.88  &  0.866
\\
\hline

\hline
\end{tabular}
}
\caption{The Pearson Correlation Coefficients between vanilla(OpenAI) model and STPEV(Mean) model.}
\label{table: Pearson Vanilla}
\end{table*}

\begin{table*}[ht]
\centering
\resizebox{\textwidth}{!}{%
\begin{tabular}{|c|c c c c c|c c c c c|c c c c c|c c c c c|c|}
\hline
\multirow{2}{*}{Year} &
\multicolumn{5}{c|}{First Quarter} &
\multicolumn{5}{c|}{second Quarter} &
\multicolumn{5}{c|}{Third Quarter} &
\multicolumn{5}{c|}{Fourth Quarter} &
\multirow{2}{*}{Yearly Average}
\\
& $\overline{Coef}$ & $Coef_3$ & $Coef_7$ & $Coef_{15}$ & $Coef_{30}$ 
& $\overline{Coef}$ & $Coef_3$ & $Coef_7$ & $Coef_{15}$ & $Coef_{30}$
& $\overline{Coef}$ & $Coef_3$ & $Coef_7$ & $Coef_{15}$ & $Coef_{30}$ 
& $\overline{Coef}$ & $Coef_3$ & $Coef_7$ & $Coef_{15}$ & $Coef_{30}$  & \\

\hline
2019 & - & - & - & - & -  &  0.113 & 0.128 & 0.119 & 0.091 & 0.114 & -0.003 & -0.076 & 0.092 & 0.022 & -0.052 & 0.369 & 0.421 & 0.405 & 0.329 & 0.319  &  0.494
\\
\hline
2020 & 0.494 & 0.566 & 0.508 & 0.451 & 0.45 & 0.372 & 0.525 & 0.382 & 0.382 & 0.198 & 0.439 & 0.542 & 0.447 & 0.39 & 0.378 & 0.579 & 0.603 & 0.567 & 0.577 & 0.57  &  0.628
\\
\hline
2021 & 0.628 & 0.616 & 0.659 & 0.618 & 0.62 & 0.707 & 0.706 & 0.702 & 0.717 & 0.704 & 0.737 & 0.733 & 0.731 & 0.753 & 0.729 & 0.732 & 0.719 & 0.741 & 0.744 & 0.724  &  0.753
\\
\hline
2022 & 0.753 & 0.73 & 0.754 & 0.754 & 0.772 & 0.709 & 0.682 & 0.714 & 0.717 & 0.72 & 0.755 & 0.719 & 0.764 & 0.77 & 0.768 & 0.799 & 0.764 & 0.794 & 0.812 & 0.824  &  0.825
\\
\hline
2023 & 0.825 & 0.805 & 0.825 & 0.834 & 0.838 & 0.817 & 0.809 & 0.822 & 0.821 & 0.815 & 0.816 & 0.79 & 0.818 & 0.831 & 0.825 & 0.81 & 0.816 & 0.801 & 0.813 & 0.81  &  0.81
\\
\hline

\hline
\end{tabular}
}
\caption{The Pearson Correlation Coefficients between vanilla(Gecko) model and STPEV(Mean) model.}
\label{table: Pearson Vanilla(Gecko)}
\end{table*}

\begin{table*}[ht]
\centering
\resizebox{\textwidth}{!}{%
\begin{tabular}{|c|c c c c c|c c c c c|c c c c c|c c c c c|c|}
\hline
\multirow{2}{*}{Year} &
\multicolumn{5}{c|}{First Quarter} &
\multicolumn{5}{c|}{second Quarter} &
\multicolumn{5}{c|}{Third Quarter} &
\multicolumn{5}{c|}{Fourth Quarter} &
\multirow{2}{*}{Yearly Average}
\\
& $\overline{Coef}$ & $Coef_3$ & $Coef_7$ & $Coef_{15}$ & $Coef_{30}$ 
& $\overline{Coef}$ & $Coef_3$ & $Coef_7$ & $Coef_{15}$ & $Coef_{30}$
& $\overline{Coef}$ & $Coef_3$ & $Coef_7$ & $Coef_{15}$ & $Coef_{30}$ 
& $\overline{Coef}$ & $Coef_3$ & $Coef_7$ & $Coef_{15}$ & $Coef_{30}$  & \\

\hline
2019 & - & - & - & - & -  &  0.258 & 0.352 & 0.23 & 0.239 & 0.213 & 0.535 & 0.594 & 0.509 & 0.525 & 0.512 & 0.555 & 0.593 & 0.539 & 0.562 & 0.528  &  0.683
\\
\hline
2020 & 0.683 & 0.704 & 0.682 & 0.69 & 0.656 & 0.654 & 0.72 & 0.671 & 0.667 & 0.558 & 0.605 & 0.655 & 0.578 & 0.571 & 0.615 & 0.693 & 0.729 & 0.689 & 0.678 & 0.678  &  0.773
\\
\hline
2021 & 0.773 & 0.775 & 0.779 & 0.76 & 0.778 & 0.752 & 0.782 & 0.762 & 0.741 & 0.725 & 0.843 & 0.879 & 0.842 & 0.835 & 0.817 & 0.863 & 0.848 & 0.889 & 0.865 & 0.852  &  0.799
\\
\hline
2022 & 0.799 & 0.807 & 0.796 & 0.798 & 0.796 & 0.781 & 0.771 & 0.771 & 0.788 & 0.796 & 0.822 & 0.812 & 0.819 & 0.842 & 0.817 & 0.883 & 0.865 & 0.883 & 0.894 & 0.892  &  0.792
\\
\hline
2023 & 0.792 & 0.803 & 0.836 & 0.833 & 0.695 & 0.842 & 0.828 & 0.838 & 0.852 & 0.847 & 0.816 & 0.764 & 0.839 & 0.832 & 0.828 & 0.848 & 0.772 & 0.861 & 0.876 & 0.884  &  0.848
\\
\hline

\end{tabular}
}
\caption{The Pearson Correlation Coefficients between GPT-4o(Summarization) model and STPEV(Mean).}
\label{table: Pearson GPT-4o(Summarization)}
\end{table*}

\begin{table*}[ht]
\centering
\resizebox{\textwidth}{!}{%
\begin{tabular}{|c|c c c c c|c c c c c|c c c c c|c c c c c|c|}
\hline
\multirow{2}{*}{Year} &
\multicolumn{5}{c|}{First Quarter} &
\multicolumn{5}{c|}{second Quarter} &
\multicolumn{5}{c|}{Third Quarter} &
\multicolumn{5}{c|}{Fourth Quarter} &
\multirow{2}{*}{Yearly Average}
\\
& $\overline{Coef}$ & $Coef_3$ & $Coef_7$ & $Coef_{15}$ & $Coef_{30}$ 
& $\overline{Coef}$ & $Coef_3$ & $Coef_7$ & $Coef_{15}$ & $Coef_{30}$
& $\overline{Coef}$ & $Coef_3$ & $Coef_7$ & $Coef_{15}$ & $Coef_{30}$ 
& $\overline{Coef}$ & $Coef_3$ & $Coef_7$ & $Coef_{15}$ & $Coef_{30}$  & \\

\hline
2019 & - & - & - & - & -  &  0.259 & 0.326 & 0.245 & 0.217 & 0.247 & 0.507 & 0.51 & 0.497 & 0.511 & 0.51 & 0.453 & 0.484 & 0.422 & 0.477 & 0.429  &  0.711
\\
\hline
2020 & 0.711 & 0.708 & 0.714 & 0.716 & 0.707 & 0.668 & 0.729 & 0.698 & 0.669 & 0.575 & 0.638 & 0.694 & 0.628 & 0.606 & 0.625 & 0.731 & 0.707 & 0.732 & 0.752 & 0.731  &  0.777
\\
\hline
2021 & 0.777 & 0.792 & 0.788 & 0.753 & 0.777 & 0.86 & 0.86 & 0.855 & 0.871 & 0.854 & 0.833 & 0.839 & 0.82 & 0.829 & 0.844 & 0.821 & 0.831 & 0.839 & 0.816 & 0.798  &  0.831
\\
\hline
2022 & 0.831 & 0.84 & 0.843 & 0.829 & 0.813 & 0.87 & 0.861 & 0.851 & 0.898 & 0.871 & 0.887 & 0.853 & 0.899 & 0.901 & 0.897 & 0.893 & 0.876 & 0.893 & 0.895 & 0.909  &  0.855
\\
\hline
2023 & 0.855 & 0.834 & 0.861 & 0.871 & 0.853 & 0.875 & 0.857 & 0.885 & 0.884 & 0.876 & 0.892 & 0.878 & 0.883 & 0.901 & 0.904 & 0.916 & 0.895 & 0.919 & 0.919 & 0.93  &  0.916
\\
\hline

\end{tabular}
}
\caption{The Pearson Correlation Coefficients between GPT-4o(Task-Specific) model and STPEV(Mean).}
\label{table: Pearson GPT-4o(Task-Specific)}
\end{table*}

\begin{table*}[ht]
\centering
\resizebox{\textwidth}{!}{%
\begin{tabular}{|c|c c c c c|c c c c c|c c c c c|c c c c c|c|}
\hline
\multirow{2}{*}{Year} &
\multicolumn{5}{c|}{First Quarter} &
\multicolumn{5}{c|}{second Quarter} &
\multicolumn{5}{c|}{Third Quarter} &
\multicolumn{5}{c|}{Fourth Quarter} &
\multirow{2}{*}{Yearly Average}
\\
& $\overline{Coef}$ & $Coef_3$ & $Coef_7$ & $Coef_{15}$ & $Coef_{30}$ 
& $\overline{Coef}$ & $Coef_3$ & $Coef_7$ & $Coef_{15}$ & $Coef_{30}$
& $\overline{Coef}$ & $Coef_3$ & $Coef_7$ & $Coef_{15}$ & $Coef_{30}$ 
& $\overline{Coef}$ & $Coef_3$ & $Coef_7$ & $Coef_{15}$ & $Coef_{30}$  & \\

\hline
2019 & - & - & - & - & -  &  0.27 & 0.409 & 0.262 & 0.211 & 0.197 & 0.598 & 0.639 & 0.6 & 0.579 & 0.575 & 0.507 & 0.522 & 0.469 & 0.53 & 0.509  &  0.726
\\
\hline
2020 & 0.726 & 0.739 & 0.729 & 0.722 & 0.714 & 0.698 & 0.735 & 0.713 & 0.7 & 0.644 & 0.669 & 0.701 & 0.669 & 0.637 & 0.672 & 0.741 & 0.759 & 0.723 & 0.755 & 0.728  &  0.837
\\
\hline
2021 & 0.837 & 0.827 & 0.849 & 0.843 & 0.831 & 0.851 & 0.845 & 0.859 & 0.852 & 0.848 & 0.872 & 0.872 & 0.86 & 0.898 & 0.859 & 0.843 & 0.836 & 0.859 & 0.86 & 0.816  &  0.852
\\
\hline
2022 & 0.852 & 0.876 & 0.869 & 0.854 & 0.808 & 0.853 & 0.854 & 0.836 & 0.877 & 0.844 & 0.838 & 0.809 & 0.844 & 0.864 & 0.837 & 0.898 & 0.884 & 0.902 & 0.911 & 0.893  &  0.854
\\
\hline
2023 & 0.854 & 0.844 & 0.859 & 0.863 & 0.851 & 0.866 & 0.845 & 0.857 & 0.867 & 0.895 & 0.845 & 0.84 & 0.825 & 0.85 & 0.864 & 0.867 & 0.858 & 0.843 & 0.88 & 0.888  &  0.867
\\
\hline

\end{tabular}
}
\caption{The Pearson Correlation Coefficients between Gemini(Summarization) model and STPEV(Mean).}
\label{table: Pearson Gemini(Summarization)}
\end{table*}

\begin{table*}[ht]
\centering
\resizebox{\textwidth}{!}{%
\begin{tabular}{|c|c c c c c|c c c c c|c c c c c|c c c c c|c|}
\hline
\multirow{2}{*}{Year} &
\multicolumn{5}{c|}{First Quarter} &
\multicolumn{5}{c|}{second Quarter} &
\multicolumn{5}{c|}{Third Quarter} &
\multicolumn{5}{c|}{Fourth Quarter} &
\multirow{2}{*}{Yearly Average}
\\
& $\overline{Coef}$ & $Coef_3$ & $Coef_7$ & $Coef_{15}$ & $Coef_{30}$ 
& $\overline{Coef}$ & $Coef_3$ & $Coef_7$ & $Coef_{15}$ & $Coef_{30}$
& $\overline{Coef}$ & $Coef_3$ & $Coef_7$ & $Coef_{15}$ & $Coef_{30}$ 
& $\overline{Coef}$ & $Coef_3$ & $Coef_7$ & $Coef_{15}$ & $Coef_{30}$  & \\

\hline
2019 & - & - & - & - & -  &  0.216 & 0.22 & 0.202 & 0.207 & 0.237 & 0.469 & 0.574 & 0.496 & 0.423 & 0.383 & 0.494 & 0.494 & 0.492 & 0.52 & 0.471  &  0.603
\\
\hline
2020 & 0.603 & 0.595 & 0.592 & 0.606 & 0.62 & 0.692 & 0.763 & 0.725 & 0.669 & 0.612 & 0.575 & 0.567 & 0.594 & 0.563 & 0.577 & 0.728 & 0.73 & 0.719 & 0.737 & 0.724  &  0.768
\\
\hline
2021 & 0.768 & 0.78 & 0.799 & 0.743 & 0.749 & 0.785 & 0.805 & 0.816 & 0.781 & 0.739 & 0.789 & 0.83 & 0.772 & 0.786 & 0.769 & 0.791 & 0.771 & 0.808 & 0.804 & 0.783  &  0.816
\\
\hline
2022 & 0.816 & 0.818 & 0.814 & 0.835 & 0.795 & 0.862 & 0.821 & 0.882 & 0.879 & 0.868 & 0.821 & 0.78 & 0.831 & 0.846 & 0.829 & 0.878 & 0.853 & 0.888 & 0.891 & 0.881  &  0.843
\\
\hline
2023 & 0.843 & 0.826 & 0.865 & 0.855 & 0.824 & 0.867 & 0.828 & 0.874 & 0.88 & 0.883 & 0.845 & 0.827 & 0.838 & 0.856 & 0.858 & 0.867 & 0.858 & 0.861 & 0.872 & 0.877  &  0.867
\\
\hline

\end{tabular}
}
\caption{The Pearson Correlation Coefficients between Gemini(Task-Specific) model and STPEV(Mean).}
\label{table: Pearson Gemini(Task-Specific)}
\end{table*}

\begin{table*}[ht]
\centering
\resizebox{\textwidth}{!}{%
\begin{tabular}{|c|c c c c c|c c c c c|c c c c c|c c c c c|c|}
\hline
\multirow{2}{*}{Year} &
\multicolumn{5}{c|}{First Quarter} &
\multicolumn{5}{c|}{second Quarter} &
\multicolumn{5}{c|}{Third Quarter} &
\multicolumn{5}{c|}{Fourth Quarter} &
\multirow{2}{*}{Yearly Average}
\\
& $\overline{Coef}$ & $Coef_3$ & $Coef_7$ & $Coef_{15}$ & $Coef_{30}$ 
& $\overline{Coef}$ & $Coef_3$ & $Coef_7$ & $Coef_{15}$ & $Coef_{30}$
& $\overline{Coef}$ & $Coef_3$ & $Coef_7$ & $Coef_{15}$ & $Coef_{30}$ 
& $\overline{Coef}$ & $Coef_3$ & $Coef_7$ & $Coef_{15}$ & $Coef_{30}$  & \\

\hline
2019 & - & - & - & - & -  &  0.101 & 0.064 & 0.11 & 0.113 & 0.118 & 0.025 & 0.014 & 0.054 & 0.03 & 0.003 & -0.116 & -0.018 & -0.073 & -0.142 & -0.23  &  -0.162
\\
\hline
2020 & -0.162 & -0.158 & -0.171 & -0.17 & -0.149 & -0.014 & -0.045 & -0.043 & 0.019 & 0.013 & 0.102 & 0.086 & 0.127 & 0.118 & 0.075 & -0.124 & -0.139 & -0.138 & -0.121 & -0.097  &  -0.082
\\
\hline
2021 & -0.082 & -0.03 & -0.078 & -0.109 & -0.11 & 0.236 & 0.163 & 0.254 & 0.247 & 0.278 & -0.106 & -0.11 & -0.082 & -0.08 & -0.149 & 0.119 & 0.069 & 0.111 & 0.173 & 0.124  &  0.183
\\
\hline
2022 & 0.183 & 0.172 & 0.19 & 0.179 & 0.192 & 0.059 & 0.088 & 0.044 & 0.056 & 0.047 & 0.111 & 0.116 & 0.094 & 0.114 & 0.122 & -0.124 & -0.072 & -0.125 & -0.134 & -0.165  &  0.007
\\
\hline
2023 & 0.007 & -0.111 & 0.026 & 0.036 & 0.075 & 0.096 & 0.065 & 0.142 & 0.092 & 0.084 & -0.146 & -0.091 & -0.166 & -0.169 & -0.156 & 0.004 & 0.052 & -0.019 & 0.006 & -0.024  &  0.004
\\
\hline

\end{tabular}
}
\caption{The Pearson Correlation Coefficients between Random(All) model and STPEV(Mean).}
\label{table: Pearson Random(All)}
\end{table*}

\begin{table*}[ht]
\centering
\resizebox{\textwidth}{!}{%
\begin{tabular}{|c|c c c c c|c c c c c|c c c c c|c c c c c|c|}
\hline
\multirow{2}{*}{Year} &
\multicolumn{5}{c|}{First Quarter} &
\multicolumn{5}{c|}{second Quarter} &
\multicolumn{5}{c|}{Third Quarter} &
\multicolumn{5}{c|}{Fourth Quarter} &
\multirow{2}{*}{Yearly Average}
\\
& $\overline{Coef}$ & $Coef_3$ & $Coef_7$ & $Coef_{15}$ & $Coef_{30}$ 
& $\overline{Coef}$ & $Coef_3$ & $Coef_7$ & $Coef_{15}$ & $Coef_{30}$
& $\overline{Coef}$ & $Coef_3$ & $Coef_7$ & $Coef_{15}$ & $Coef_{30}$ 
& $\overline{Coef}$ & $Coef_3$ & $Coef_7$ & $Coef_{15}$ & $Coef_{30}$  & \\
\hline
2019 & - & - & - & - & -  &  -0.071 & -0.014 & -0.099 & -0.068 & -0.103 & 0.055 & 0.096 & 0.069 & 0.041 & 0.015 & 0.222 & 0.367 & 0.223 & 0.175 & 0.125  &  0.169
\\
\hline
2020 & 0.169 & 0.253 & 0.154 & 0.138 & 0.133 & 0.316 & 0.499 & 0.358 & 0.219 & 0.189 & 0.308 & 0.379 & 0.342 & 0.273 & 0.238 & 0.506 & 0.606 & 0.541 & 0.481 & 0.395  &  0.47
\\
\hline
2021 & 0.47 & 0.637 & 0.657 & 0.318 & 0.267 & 0.587 & 0.637 & 0.617 & 0.575 & 0.519 & 0.641 & 0.624 & 0.706 & 0.665 & 0.569 & 0.726 & 0.765 & 0.749 & 0.71 & 0.682  &  0.786
\\
\hline
2022 & 0.786 & 0.913 & 0.788 & 0.753 & 0.691 & 0.878 & 0.885 & 0.89 & 0.879 & 0.859 & 0.912 & 0.896 & 0.92 & 0.927 & 0.907 & 0.898 & 0.905 & 0.877 & 0.912 & 0.899  &  0.925
\\
\hline
2023 & 0.925 & 0.885 & 0.946 & 0.932 & 0.938 & 0.949 & 0.929 & 0.955 & 0.96 & 0.951 & 0.916 & 0.888 & 0.92 & 0.926 & 0.931 & 0.946 & 0.939 & 0.952 & 0.952 & 0.94  &  0.946
\\
\hline
\end{tabular}
}
\caption{The Pearson Correlation Coefficients between Random(Ticker) model and STPEV(Mean).}
\label{table: Pearson Random(Ticker)}
\end{table*}

\end{document}